\documentclass[10pt,journal,compsoc]{IEEEtran}

\ifCLASSOPTIONcompsoc
\usepackage[nocompress]{cite}
\else
  \usepackage{cite}
\fi

\ifCLASSINFOpdf
  \usepackage[pdftex]{graphicx}
  \graphicspath{{../pdf/}{../jpeg/}}
  \DeclareGraphicsExtensions{.pdf,.jpeg,.png}
\else
\fi

\usepackage{microtype}

\usepackage{hyperref}
\hypersetup{pdftex,
   pagebackref=true,
   pdfpagemode=UseNone,
   pdftitle={\@title},
   pdfauthor={\@author},
   pdfsubject={},
   pdfkeywords={},
   pageanchor=true,
   plainpages=false,       % for problems with page referencing
   hypertexnames=false,    % for handling subfigures correctly
   bookmarks=true,         % we want bookmarks
   bookmarksnumbered=false,% include the section numbers in the list
   bookmarksopen=true,     % In the list, display highest level only
   bookmarksopenlevel=3,   % display three levels of bookmarks
   pdfpagemode=UseNone,    % show just the page
   pdfstartview=Fit,       % defail page view is the whole page at once
   pdfborder={0 0 0},      % we don't want those silly boxes for links
   breaklinks=true,        % allow breaking links
   colorlinks=false,       % don't change coloring of links
   nesting=true,
   linktocpage}
\usepackage[T1]{fontenc}% use always vector fonts if available

\usepackage{mathptmx}
\usepackage{times}
\usepackage{subfigure}
\usepackage[table,xcdraw]{xcolor}
\usepackage{enumitem}
\usepackage{textcomp}
\usepackage{expdlist}
\usepackage[most]{tcolorbox}
\usepackage[utf8]{inputenc}
\usepackage{multicol}       % needed for the additional material typesetting
\usepackage{dblfloatfix}    % To enable figures at the bottom of page

\usepackage{subfiles} % Best loaded last in the preamble

\usepackage{comment}

\newcommand{\eg}{e.\,g.}
\newcommand{\ie}{i.\,e.}

\definecolor{tobiascolor}{RGB}{14, 149, 229}
\colorlet{bobcolor}{green!5!orange!95!}
\definecolor{michaelcolor}{RGB}{1,129,1}
\definecolor{hanweicolor}{RGB}{20, 20, 140}
\definecolor{torstencolor}{RGB}{149, 229, 7}
\definecolor{jiancolor}{RGB}{20,139,20}

\newcommand{\changes}[1]{\textcolor{blue}{#1}}
\renewcommand{\changes}[1]{#1}
\newcommand{\changesR}[1]{\textcolor{blue}{#1}}
\renewcommand{\changesR}[1]{#1}

\usepackage{chngcntr}

\ifCLASSINFOpdf
\else
\fi

\usepackage{xspace}
\newcommand{\dataname}{VIS30K\xspace}
\newcommand{\toolname}{VISImageNavigator\xspace}
\newcommand{\toolnameshort}{VIN\xspace}
\newcommand{\imagesall}{29,689\xspace}

\newcommand{\imagesfig}{26,776\xspace}
\newcommand{\imagestab}{2,913\xspace}
\newcommand{\imagesvis}{13,509\xspace}
\newcommand{\imagesscivis}{3,232\xspace}
\newcommand{\imagesinfovis}{7,834\xspace}
\newcommand{\imagesvast}{5,114\xspace}
\newcommand{\papers}{2,916\xspace}

\begin{document}

\title{\dataname: A Collection of Figures and Tables from {IEEE} Visualization Conference Publications}

% author names and affiliations
% use a multiple column layout for up to three different
% affiliations
\author{\IEEEauthorblockN{Jian~Chen, Meng Ling, Rui Li, Petra~Isenberg, Tobias~Isenberg, Michael Sedlmair,  Torsten~M{\"o}ller, Robert~S.~Laramee, Han-Wei~Shen, Katharina W{\"u}nsche, and Qiru Wang
}
\IEEEcompsocitemizethanks{%
%\IEEEcompsocthanksitem M. Ling, J. Chen, R. Li, H. Shen are with the Computer Science and Engineering, Ohio State University, OH 43210, USA.
\IEEEcompsocthanksitem J. Chen, M. Ling, R. Li, and H.-W. Shen are with %the Computer Science and Engineering, 
The Ohio State University, USA.
%\protect\\
% note need leading \protect in front of \\ to get a newline within \thanks as
% \\ is fragile and will error, could use \hfil\break instead.
E-mails: \{chen.8028\,$|$\,ling.253\,$|$\, li.8950\,$|$\,shen.94\}@osu.edu.
\IEEEcompsocthanksitem P. Isenberg and T. Isenberg are with Inria, France. 
%\IEEEcompsocthanksitem P. Isenberg and T. Isenberg are with Inria, 91190 Gif-sur-Yvette, France. 
%\protect\\
E-mails: \{petra.isenberg\,$|$\, tobias.isenberg\}@inria.fr.
\IEEEcompsocthanksitem M. Sedlmair is with University of Stuttgart, Germany. 
%\IEEEcompsocthanksitem M. Sedlmair is with University of Stuttgart, 70174 Stuttgart, Germany. 
%\protect\\
E-mail: michael.sedlmair@visus.uni-stuttgart.de.
\IEEEcompsocthanksitem T. M{\"o}ller and K. W{\"u}nsche are with University of Vienna, Austria.
%\IEEEcompsocthanksitem T. M{\"o}ller and K. W{\"u}nsche are with University of Vienna, Wien 1010, Austria.
%\protect\\
E-mails: \{torsten.moeller $|$ katharina.wuensche\}@univie.ac.at.
\IEEEcompsocthanksitem R. S.\ Laramee and Q.\ Wang are with the University of Nottingham, UK. 
%\IEEEcompsocthanksitem R. S. Laramee is with Swansea University, Swansea SA2 8PP, UK. 
%\protect\\
E-mail:
\{robert.laramee\,$|$\,qiru.wang\}@nottingham.ac.uk.
%\IEEEcompsocthanksitem H.-W. Shen is with Ohio State University, USA. 
%\IEEEcompsocthanksitem R. S. Laramee is with Swansea University, Swansea SA2 8PP, UK. 
%\protect\\
%E-mail: shen.94@osu.edu.
% <-this % stops an unwanted space
%\thanks{Manuscript received April 19, 2005; revised August 26, 2015.}
}
}

\markboth{submitted to IEEE Transactions on Visualization and Computer Graphics}%
{Ling \MakeLowercase{\textit{et al.}}: VIS29K}

\IEEEtitleabstractindextext{%
\begin{abstract} 
%Thirty years (1990-2019) of IEEE Visualization publications have brought us over 30K figures. 
%we make this brilliant world of published figures available and use convolutional neural network (CNN) models to extract them. 
We present the \dataname dataset, a collection of \imagesall images that represents 30 years of figures and tables from each track of the IEEE Visualization conference series (Vis, SciVis, InfoVis, VAST). \dataname's comprehensive coverage of the scientific literature in visualization not only reflects the progress of the field but also enables researchers to study the evolution of the state-of-the-art and to find relevant work based on graphical content.
We describe the dataset and our semi-automatic collection process, which couples convolutional neural networks (CNN) with curation. Extracting figures and tables semi-automatically allows us to verify that no images are overlooked or extracted erroneously. 
To improve quality further, we engaged in a peer-search process for high-quality figures from early IEEE Visualization papers. With the resulting data, we also contribute \toolname (\toolnameshort, \href{https://visimagenavigator.github.io/}{\texttt{visimagenavigator.github.io}}), a web-based tool that facilitates searching and exploring \dataname
by author names, paper keywords, title and abstract, and years.
\end{abstract}
%also offers information retrieval through figures and tables. 
%our carefully curated figure and table collection, 

% Note that keywords are not normally used for peerreview papers.
}

%customized for visualization publications to automatically extract figures with minimal human intervention. 
%We further present a few heuristics and called the process \$2.  

%The key challenge in analyzing the visualization literature is the extraction of the figures and tables. We address this challenge by

\maketitle

% As a general rule, do not put math, special symbols or citations
% in the abstract

% For peer review papers, you can put extra information on the cover
% page as needed:
% \ifCLASSOPTIONpeerreview
% \begin{center} \bfseries EDICS Category: 3-BBND \end{center}
% \fi
%
% For peerreview papers, this IEEEtran command inserts a page break and
% creates the second title. It will be ignored for other modes.
\IEEEpeerreviewmaketitle

%about dataport
% https://www.ilovephd.com/how-to-access-free-ieee-datasets/

\IEEEraisesectionheading{\section{Introduction}\label{sec:introduction}}

%\section{Introduction}
%\label{sec:introduction}

\IEEEPARstart{V}{isualization} is a discipline that inherently relies on images and videos to explain and showcase its research.  Images are thus an essential component of scientific publications in our field. They facilitate comprehension of complex scientific concepts~\cite{firat2018towards,strobelt2009document} and enable authors to refer to their proposed visualization solutions, alternatives, and competing approaches or to graphically explain algorithms, techniques, workflows, and study results.

Browsing a domain's images can reveal temporal trends and common practices.
%\add{~\cite{strobelt2009document}. 
It facilitates the comparison of sub-disciplines~\cite{lee2017viziometrics}. Although figures are ubiquitous in visualization publications, they are embedded in PDFs and 
%often reside behind a paywall.  Hence, these images
remain largely inaccessible via scholarly search tools such as digital libraries, Google Scholar, CiteSeerX, or MS Academic Search.
%, except Semantic Scholar. 
%\ms{do we  really want to make the paywall argument here? In the end, if we put the data onto IEEE it might be behind the same paywall again, right?} 
The primary goal of our work---similar to that of past work on IEEE VIS papers \cite{isenberg2016vispubdata}, keywords \cite{isenberg2016visualization}, or EuroVis papers \cite{Smith:2020:IEP}---is to extend the corpora of data we can use for studies of the visualization field. 
%We contribute a large, diverse, and high-quality visualization image database that can serve as the basis of applying deep learning algorithms to solve visualization problems. Akin to data collections in computer vision, these images themselves already contain well-defined knowledge. 
%

\begin{figure*}[!t]
\centering
\includegraphics[width=\textwidth]{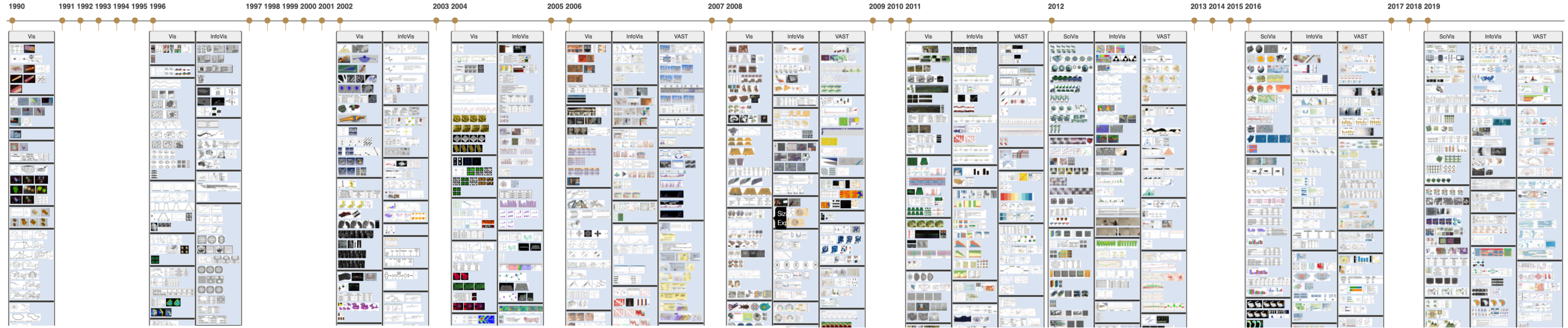}\vspace{-1ex}
\caption{\protect\changes{A timeline of selected images from all 30 years (1990---2019) of IEEE Visualization conference showing diverse and trending research work. \changesR{Best viewed electronically, zoomed in.}}}%\vspace{-2ex}
\label{fig:vis30k-timeline}
\end{figure*}

%JC added 
Our primary contribution is a dataset we call \dataname (\autoref{fig:vis30k-timeline}). %and~\ref{fig:allYearsCount}). 
It contains images and tables from 30 years (1990--2019) of the IEEE VIS conference, spanning all tracks: Vis, InfoVis, SciVis, and VAST.
IEEE VIS is the longest-running and largest conference focusing on visualization and its images reflect the evolution of the field. %than those at younger venues or 
\changes{Our primary data sources include IEEE Xplore, conference CDs, and hard copies of the conference proceedings from which we obtain the images in their best possible quality.}
%those with mixed content 
%\todo[inline]{TM: what do you mean here with 'mixed content'? This is unclear.} %such as 
%broader graphics and HCI events. 
In addition to images, we include tables as special form of data organization that can be 
%\add{(e.g., validation results are often in tables). Algorithms and pseudo-code and equations reflect the core methods}
informative to the community.
%\todo[inline]{TM: what do you mean with 'styling' here? to me this is not interesting (and I didn't see much variation going through my 1500 pages). What I find interesting is perhaps the extend of their use.} 
\changes{Our dataset can serve many purposes. It enables researchers to study the visual evolution of the field from an objective, image-centric point of view. It assists teaching about visualization by providing fast visual access to refereed research images and contributions.  It also can serve as a data source for researchers in other fields such as computer vision or machine learning. And, finally, it supports visualization researchers when browsing and discovering new work.}

% Visualization scholars now have a new way to search for related work (particularly from older years) and can also use the data in new research projects, for example to study visual vocabulary \cite{bertin1983semiology,munzner2014visualization}. Practitioners can use the images as inspiration in search for novel types of visual representations.

%\remove{
Collecting these figures and tables was challenging. \changes{%
We optimized data quality with a hybrid solution.
%of DeepPaperComposer~\cite{ling2020deeppapercomposer}. 
We first extracted figures via convolutional neural networks (CNNs), followed by expert curation. This way we ensure reliable data (\ie, completeness and image locations/dimensions), while at the same time requiring a manageable amount of manual cleaning and verification.%
}

Our secondary contribution is a web-based tool, \toolname (\toolnameshort, \href{https://visimagenavigator.github.io/}{\texttt{visimagenavigator.github.io}}), that \changes{allows people to search and explore our dataset}. We cross-link \toolnameshort to the metadata of KeyVis~\cite{isenberg2016visualization} and VisPubData \cite{isenberg2016vispubdata} \changes{and their detailed bibliometric metadata. This metadata associated to papers, and thus all images, allows us to support searching using text-based queries.}

\section{Related Work}
\label{sec:relatedwork}
% long version before adding the bio

Previous work from three areas inspired our own.  The first surveys past work and offers \changes{visual access to past publications.} A second group collects and analyzes metadata derived from visualization research papers. The third group relates to the CNN-based extraction algorithms we employed for our data extraction.

% 
%Previous work from IEEE VIS meta-data collection and surveys with image collections has inspired ours. 

\textbf{Visual Collections of Visualization Research.}
%\add{Data collection is the process of gathering and measuring information on variables of interest, in an established systematic fashion that enables one to answer research questions, test hypotheses, and evaluate outcomes. The renewed excitement in deep neural network research in images has dramatically increased in the past year. Part of this excitement stems from the belief that large data collections are a leap towards human-machine teaming ways of discovery.}
\changes{We are not the first to attempt a visual overview of the visualization field. Yet past work generally focuses on specific subareas of the research, each of which provides an overview of work on the subtopic rather than a comprehensive and browsable image database.}
%and published online image collections related to the respective theme. 
For instance, some work has focused on providing references and representative images of specific data or layout techniques:  Schulz's 300 tree-layout methods~\cite{schulz2011treevis}, Kucher and Kerren's more than 470 text visualizations~\cite{Kucher2015}, Aigner et al.'s over 100 temporal data visualizations~\cite{Aigner2011}, and Kehrer and Hauser's multivariate and multifaced data of more than 160 images~\cite{Kehrer2013a}.
Others examine specific visualization applications, \eg, Kerren et al.'s biological data~\cite{Kerren2017a}, Kucher et al.'s sentiment analysis~\cite{Kucher2018a}, Chatzimparmpas and Jusufi's trustworthy machine learning~\cite{Chatzimparmpas2020}, and Diehl et al.'s
VisGuides~\cite{Diehl2018} on advice and recommendations on visual design. In contrast to these focused perspectives, \dataname provides a broader coverage of all 30 years of IEEE VIS. 
\changes{Our downloadable dataset comprises all the images contained in each paper, rather than just a few samples per publication or approach}. 

\changes{The work most closely related to ours is Deng et al.'s VisImages collection of IEEE InfoVis and IEEE VAST images~\cite{Deng:2020:VLH} and Zeng et al.'s VISstory~\cite{dong2019vistory}. Both sets of authors plan to release their datasets but only provide a subset of our data. VISstory only covers data from 2009--2018, while VisImages does not include IEEE Vis and SciVis paper images. Our work also differs in its approach to quality control. We rely on expert input to check the capture of all images, while VisImages uses crowd-sourcing. VISstory only tests a subset of images for quality. Similar to these tools, we provide a web-based tool
to explore the image data although focus on different aspects.
%With both approaches we share a visual interface to view the data, although each focuses on different aspects. 
VisImages categorizes image content in addition to metadata and VISstory focuses on a paper rather than image-centered views where each paper is encoded as a ring with sectors standing for individual images.}

% \ti{Need to discuss \cite{Deng:2020:VLH}. Differences: we cover all of visualization, not only VAST and InfoVis; we did not use crowd-sourcing, and instead fully relied on expert input; we linked to existing datasets on visualization publication meta data}

%\remove{
\textbf{Meta-Analysis of Visualization Publications.}
\changes{Another direction of research centers on meta-analyses of the visualization field, without focusing on visual content.}
Lam et al.{~\cite{lam2011empirical}}, \eg, established seven empirical evaluation scenarios by analyzing 850 papers appearing in the `information visualization' subcommunity of IEEE VIS. 
%They published the full list of papers along with their classifications online. 
Isenberg et al.{~\cite{isenberg2013systematic}} later extended this historical analysis of evaluation practices to all tracks of IEEE VIS in a systematic review of 581 papers.
%also included non-empiric evaluation types. 
%Liu et al. studied four main categories of empirical methodologies, user interactions, visualization frameworks, and applications. 
%Investigating the classification of the keywords uses, 
Isenberg et al.{~\cite{isenberg2016visualization}} further collected IEEE VIS paper keywords to derive
visualization topics from a metadata collection of IEEE VIS publications{~\cite{isenberg2016vispubdata}}. We make use of metadata from this collection in our work to gather paper PDFs prior to automatic extraction.
%s used in each visualization research paper and studied how they compare to keywords enforced by the IEEE Visualization conference's paper submission system. 
%The companion results can be searched via a keyword browser 
%\texttt{\href{http://keyvis.org/}{keyvis\discretionary{}{.}{.}org}}. 
%All VIS papers are accumulated in a public dataset VisPubData.
%about visualization publications was made 
%available
%at \texttt{\href{https://vispubdata.org/}{vispubdata\discretionary{}{.}{.}org}}~\cite{isenberg2016vispubdata}.
%
Conceptually, our new \dataname extends this line of work by 
% offering more complete data collection and 
leveraging new \changes{image-based} extraction methods~\cite{ling2020deeppapercomposer} and search tools to make the figure \changes{and table} data accessible.

\textbf{CNN-based Extraction Algorithms.}
Using data-driven CNN algorithms to train classifiers to extract figures and tables is becoming increasingly popular \cite{paliwal2019tablenet,siegel2016figureseer,clark2016pdffigures}.
%become  common practice in document analysis. 
%\todo[inline]{please include a ref}
%CNNs typically require a large labeled training dataset (ground-truth) that is then used to train the classifier.
\changes{Current approaches to fine-grained recognition involve% 
%two steps: first, recruit experts to annotate a dataset of images, often in the form of part annotations and bounding boxes. Second, train a model utilizing this data to predict labels from new data. 
%Finally, fine-clean the data to obtain the ground truth. 
%\ms{it sounds like three steps. Either change two to three or the last sentence. In general, it is a bit confusing that the end result is the ground truth. Afaik, usually the labeled data is called the GT, which is then used to train a model,which then is validated against the training set, and then can be used to do things automatically. Do you mean there is a cleaning step after the prediction (last step)? Or after the gathering of human labels (first step)? }
}
the important
%preliminary
step %is
of preparing the annotated training data with ground-truth labeling prior to training a model for prediction. 
Problems in this area have inspired research in
three major directions.
%crowd sourcing, mining information from XML~\cite{clark2016pdffigures}, or manually defined rules and features.
The first is crowdsourcing to annotate the document manually~\cite{siegel2016figureseer}. 
We did not use this solution for %at least 
two reasons. 
First, cleaning noisy crowdsourced annotations is time-consuming in itself and also needs effective quality control~\cite{siegel2018extracting}. 
Second, crowdsourcing lacks flexibility: often we must know in advance if we are to extract figures, equations, texts, tables or all of these.  
%\ms{statement is a bit weird as in our case (figures, equations, abstracts, ...) that should be quite clear -- see also my general comment (will send in Email)}
%Depending on the 
%figure-extraction goal, the labels can be just figure and table (as in our case). In other analyses, they can also be the figure component parts such as axes, ticks, data, or legend, or the figure types such as bar chart, maps, vector field, etc. 
It may not be realistic to determine complete categories in advance. 
%\ms{I think the second reason is not for this paper. It seems to be a relic from the analysis paper?}
Another solution is to mine information from an XML schema~\cite{clark2016pdffigures, lopez2009grobid} 
or from \LaTeX~\cite{siegel2018extracting} or \changes{PDF~\cite{li2019figure}} syntax. 
We could not use this 
%the XML schema based solution 
approach since early IEEE VIS PDF papers %\changes{are either paper-based} or 
are 
%image-based or
lack corresponding \LaTeX\ or XML source files.
The most popular figure-extraction algorithms rely on manually defined rules and assumptions. 
%Most existing work relies on carefully engineered rules and features. 
These techniques are typically successful for the particular type of figure for which these rules are followed, but suffer from the classical problem with rule-based approaches: when rules are broken, the algorithm fails. For example, an intuitively reasonable rule is to assume captions always exist. An algorithm can locate a figure by searching for caption terms such as \emph{Fig.} and \emph{Table}~\cite{clark2016pdffigures, li2019figure}. However,  about $2\%$ of our VIS30K images do not satisfy this assumption and thus can cause the extraction algorithm to fail. 
\changes{Choudhury et al.'s algorithm~\cite{ray2015automatic} focuses on specific figure types, such as line charts and scatter plots, while our goal is to extract a comprehensive collection of figures and tables.}
For these reasons, annotated ground-truth data are not publicly available for automatic figure and table extraction.
\begin{figure*}[!t]
\centering
%\captionsetup[subfloat]{farskip=5pt,captionskip=2pt}

%\hfil
%\subfigure[\changes{A timeline view of selected images of the entire 30 years IEEE Visualization conference showing the diverse and trending research work.}]{\includegraphics[width=0.98\textwidth]{Figures/paperCard.pdf}
%}\\[.5ex]
\subfigure[Total \# of images (figures and tables), by year.]{\includegraphics[width=0.9\textwidth]{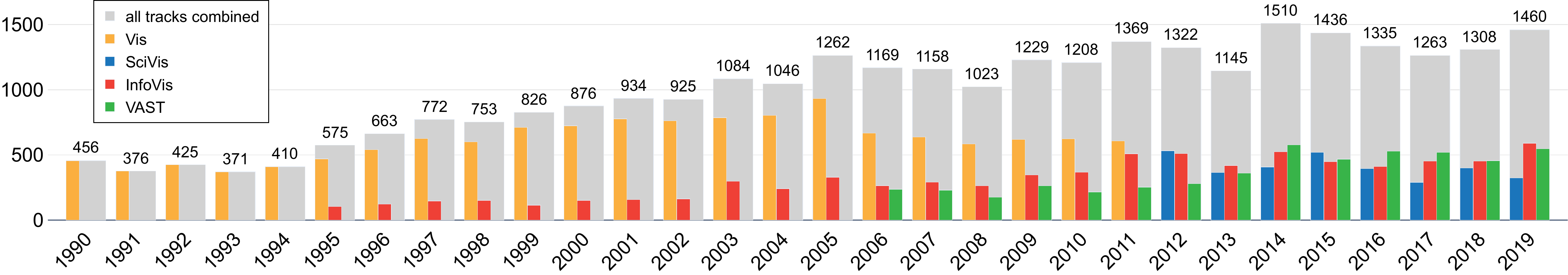}
\label{fig:byConferenceType}}\\[.5ex]
%\hfil
%\subfloat[Total \# of figures and tables vs. Year]{\includegraphics[width=\columnwidth]{Figures/imageTotalYear.png}%
%\label{fig:totalImages}}
%\hfil

%\subfloat[Average \# of images (figures and tables) per page, by year.]{\includegraphics[width=\textwidth]{Figures/averageNumImages.png}%
\subfigure[Average \# of images (figures and tables) per page, by year. \protect\changes{We included potential color plate pages from early years (1990--2001) in the page count for this analysis, and only counted an image once if that same image appeared in both the paper and its color plate.}]
{\includegraphics[width=0.9\textwidth]{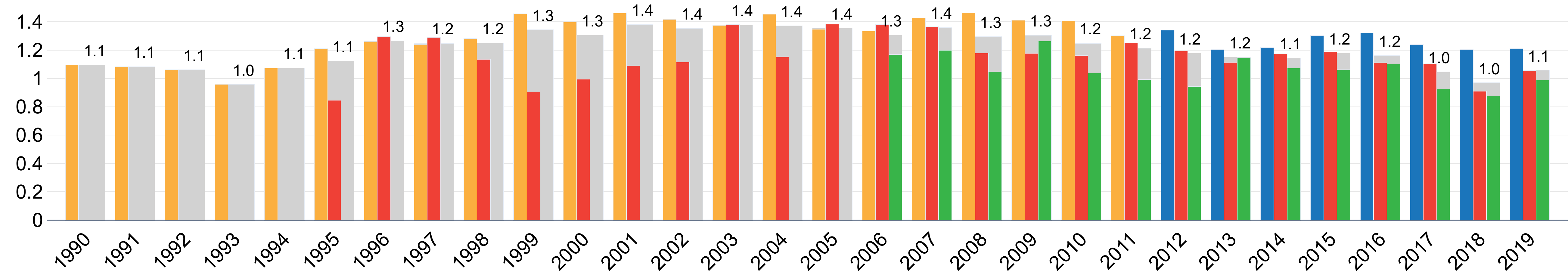}%
\label{fig:averagePerPage}}\vspace{-2ex}
\caption{%
We extracted \imagesall images (\imagesfig figures, \imagestab tables) 
%\add{\imagesalgo algorithm and pseudocode, and \imagesequ equations}
from the \papers IEEE Visualization conference papers, spanning 30 years
(Vis: \imagesvis; SciVis: \imagesscivis; InfoVis: \imagesinfovis; VAST: \imagesvast). Numbers for the joint conference are depicted as wide pale gray bars. The individual tracks are overlaid on top. 
On average, Vis/SciVis has more images per paper page than InfoVis and VAST.%
%\ms{should we split into two Figures? The caption seems only to fit for Fig1(b)...}%
}
%\ms{what about we do a stacked bar chart instead? The data is not continuous, but ordinal. But more importantly, the stacked bar chart would additionally show better the overall trend.}
%\todo[inline]{I think you should use the exact numbers here, not rounded. You also don't round the number of papers; also, the avg number of images per paper must be off by a factor of 10!} 
%}
\label{fig:allYearsCount}
\end{figure*}

%\begin{figure*}[!thp]
%\centering
%%\subfloat[CaseI]{\includegraphics[width=2.5in]{box}%
%%\label{fig_first_case}}
%%\hfil
%%\subfloat[Case II]{
%\includegraphics[width=\textwidth]{Figures/complexity.pdf}%
%%}
%\caption{The formatting of figures and tables shows great variation: here we place subfigures side-by-side for comparison to show different techniques. Subfigures can contain tabular views of different parameter choices. Figure captions sometimes appear inside the figure's rectangular bounding box. Tables often contain visual separators but the content can be hierarchical and can contain figures. The presence of these variations leads us to use composite figures and tables in our data cohort to preserve the functional values of these paper elements. \textcolor{red}{(updates needed; also the references should be in this paper's reference list, shoudn't they?)}}
%\label{fig:compositionComplexity}
%\end{figure*}

\section{Dataset Description}
\label{sec:dataset}

We now describe the data format and information stored for each figure and table in our database and the decisions we made concerning the figure extraction task. But we start by defining the terms we use throughout the remainder of the paper. 
%In the subsequent section, we will explain the collection and image/table extraction process. 

\subsection{Terms}

%\begin{description}[\compact\setlabelphantom{}]
\textbf{IEEE VIS.}  \changes{Over its history, IEEE VIS has undergone a number of name  changes (~\autoref{fig:allYearsCount}).  It started  out in 1990 as IEEE Visualization (Vis), then added IEEE InfoVis  in  1995  followed by IEEE VAST in 2006.  In 2008--2012, all three venues were jointly called IEEE VisWeek, and since 2013 the blanket name IEEE VIS has been used. From 2013 onward, the IEEE Vis ceased to exist, replaced by the IEEE Scientific Visualization (SciVis) conference.  Here we use VIS to refer to all four venues: Vis, InfoVis, VAST, and SciVis, for the entire time period covered by our dataset.}

\textbf{Figures and Tables.} We refer to a \textit{figure} as a container for graphical representations. These representations can be images, screenshots of visualization techniques, user interfaces, photos, diagrams, and others. 
\changes{We classify algorithms, pseudocode, and equations as textual content and thus do not include them in VIN.  Including these additions is left as future work.} 
%\textbf{Tables.} 
\changes{A \textit{table} is a row-column representation of relations among related data concepts or categories{~\cite{hurst2006towards}}, usually composed of cells~\cite{khusro2015methods,long2005model}.}
%\end{description}

%\todo{JC: talk about how those late breaking results, case studies, and short papers are introduced.}

\subsection{\changes{Image Data Collection}}
%\jian{change to Collected Image Data}

%{Our dataset collection includes figures and tables in the {\papers} conference and journal
%\petra{conference and journal?} 
%publications of the IEEE VIS conference from 1990 to 2019. Our collection also includes case studies and late-breaking results  from earlier years as well as short papers in recent years in order to reflect the progress of the conference.

We collected {\imagesall} images ({\imagesfig} figures and {\imagestab} tables
%\imagesalgo algorithms and pseudocode, and \imagesequ equations) 
%\todo[inline]{again, please use exact numbers
from {\papers} conference and journal
%\petra{conference and journal?} 
publications of the IEEE VIS conference from 1990 to 2019 (\autoref{fig:allYearsCount}).
%breaks these numbers down by publication year and conference tracks. 
\changes{Our collection also includes case studies and late-breaking results from earlier years as they are included in the digital library. We do not include the more recent short papers as only 3 years of data are available.  We also exclude posters as they do not appear consistently in the IEEE Xplore digital library}, our primary data source for the paper PDFs.

%This image collection and its meta-data are released through the IEEE dataport \url{http://dx.doi.org/10.21227/4hy6-vh52}}.

\changes{
We include tables as a separate category alongside figures as we consider them, unlike unstructured text, as a form of structured and visual data representation that might be useful to analyze.
Not only the visual layout of tables may be interesting, but also, more importantly, the relative frequency of tables in published research results and the amount of space tables occupy in papers. Data stored in tables can be further extracted and cross-linked into a knowledge base. Tables can be filtered out for other use cases such as searching for related work or searching for images for teaching.}

\begin{figure*}[!thp]
\begin{tcbitemize}[raster columns=3,raster equal height,colframe=gray!35, colback=white, left=0mm,right=0mm,top=0mm,bottom=0mm,boxsep=1mm]
\tcbitem
\centering
\footnotesize
\includegraphics[height=27mm]{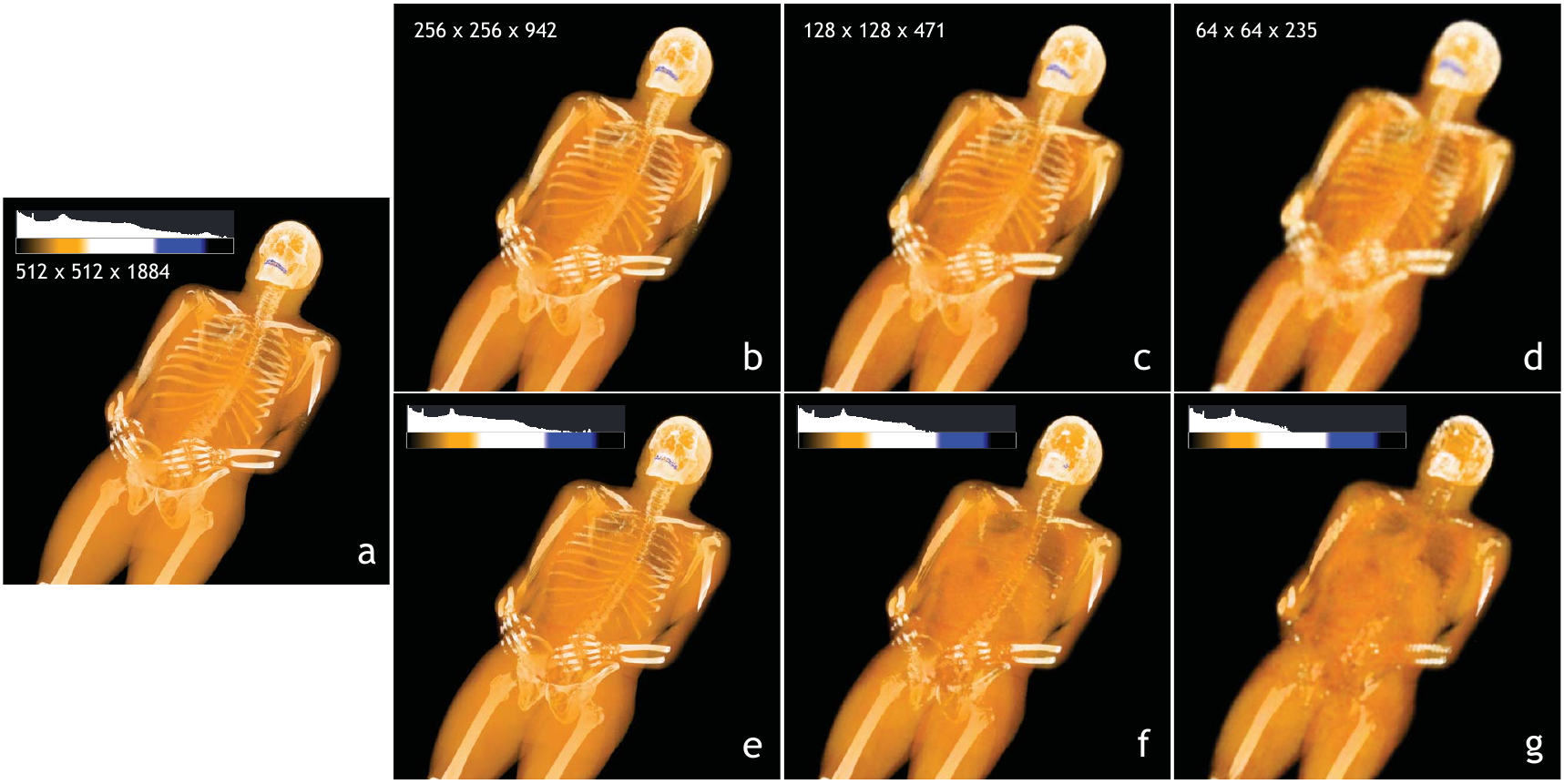}\\
(a) Sicat et al. \cite{Sicat:2014:SPV}, Fig.~7\\
Goal: comparison of volume rendering coarse levels with subcaptions embedded in the figure
\tcbitem
\centering
\footnotesize
\vspace{2mm}
\includegraphics[height=23mm]{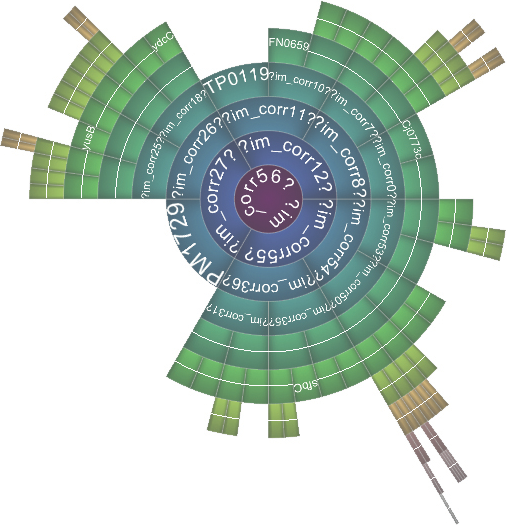}\hfill%
\includegraphics[height=24mm]{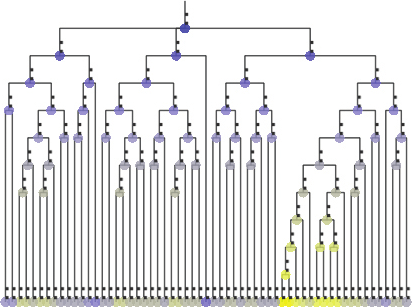}\\[1mm]
(b) Isenberg and Carpendale \cite{Isenberg:2007:ITC}, Fig.~2\\
Goal: showing two different tree layouts and labeling without subcaption
\tcbitem
\centering
\footnotesize
\vspace{1mm}
\includegraphics[height=25mm]{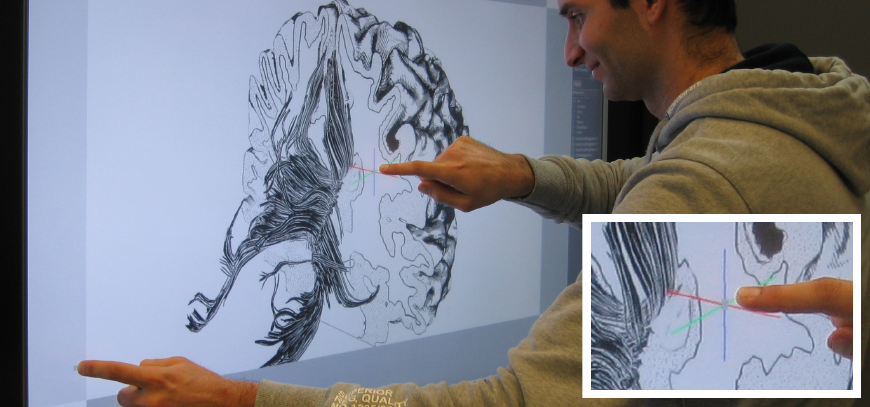}\\[1mm]
(c) Yu et al. \cite{Yu:2010:FDT}, Fig.~11\\
Goal: showing interaction technique by embedded views
\end{tcbitemize}%
\begin{tcbitemize}[raster columns=4,raster equal height,colframe=gray!35, colback=white, left=0mm,right=0mm,top=0mm,bottom=0mm,boxsep=1mm]
\tcbitem
\centering
\footnotesize
\includegraphics[height=17mm]{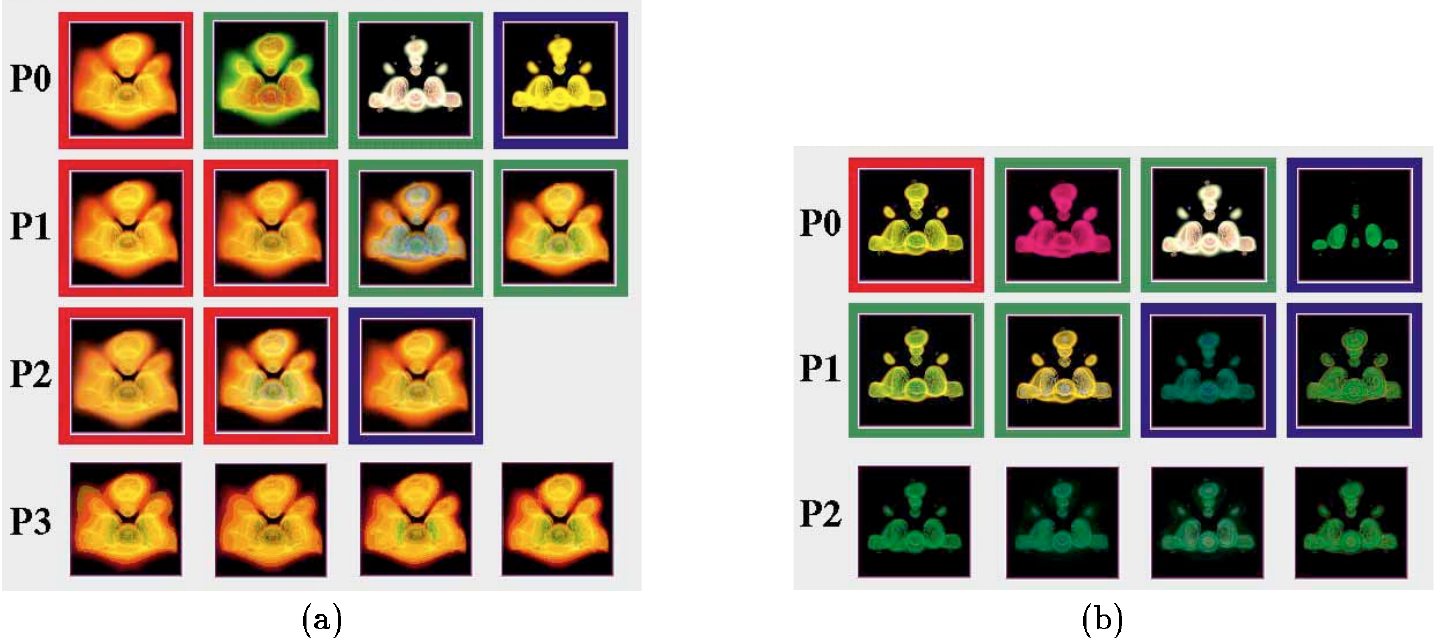}\\
(d) He et al. \cite{He:2010:GTF}, Fig.~7\\
Goal: comparing transfer function design; many views
\tcbitem
\centering
\footnotesize
\includegraphics[height=17mm]{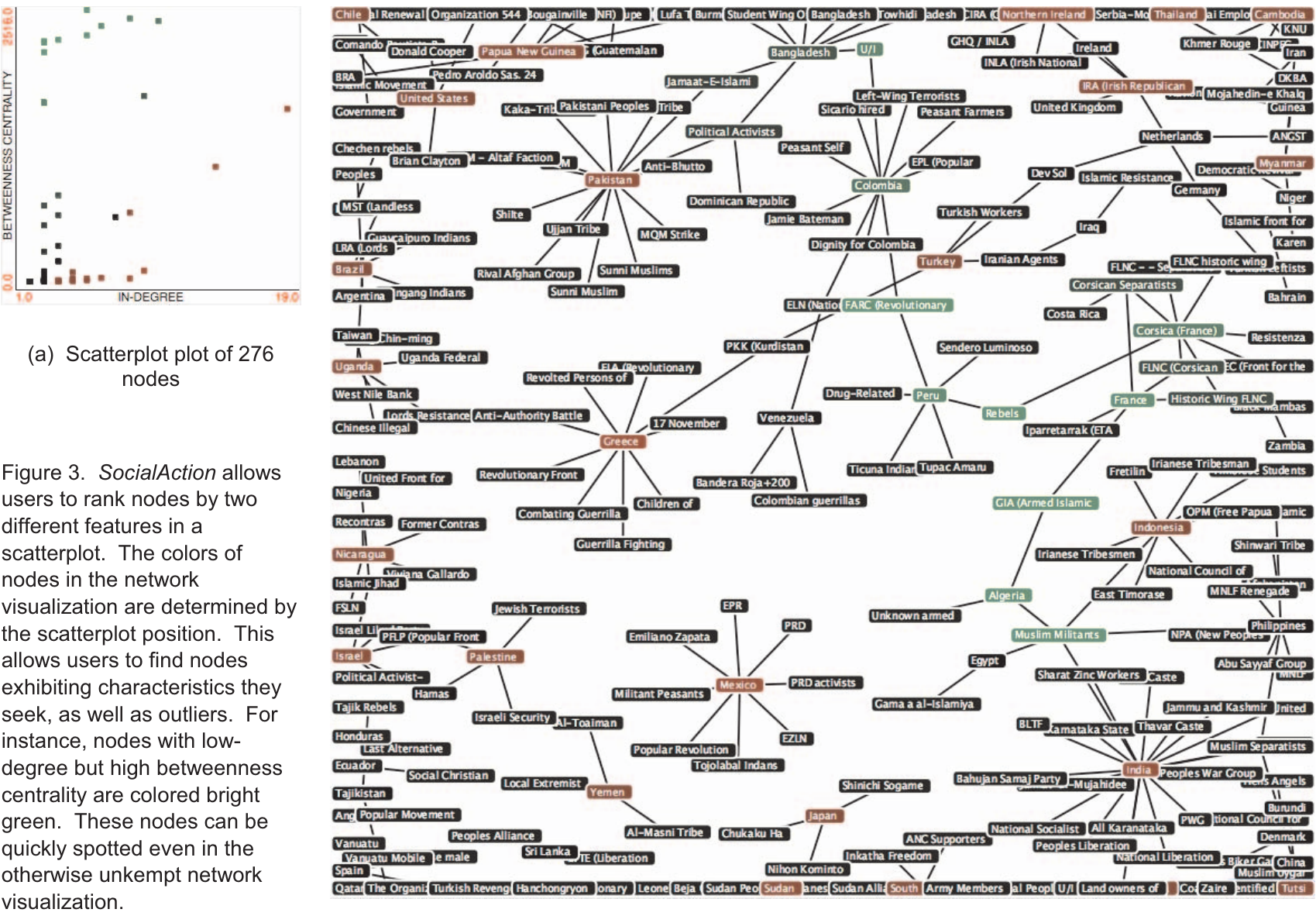}\\
\textls[-15]{(e) Perer and Shneiderman \cite{Perer:2006:BSF}, Fig.\,3}\\
Goal: saving space by placing the figure caption into an empty corner; we did not remove such captions %\textcolor{red}{???} 
%The two views support interactive brushing-and-linking. Figure caption is placed inside the interface.
\tcbitem
\centering
\footnotesize
\includegraphics[height=17mm]{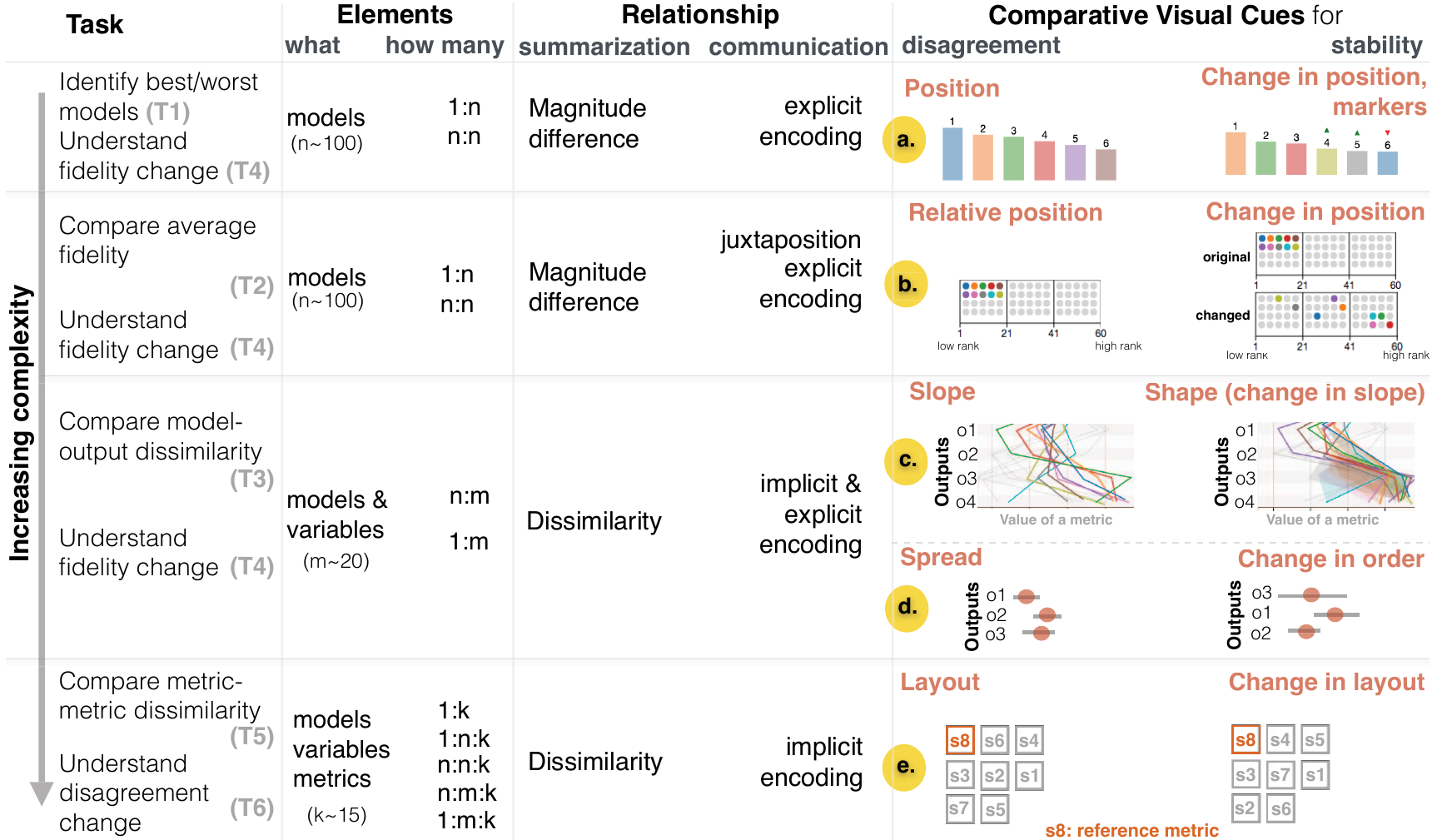}\\
(f) Dasgupta et al. \cite{Dasgupta:2020:SWC}, Fig.~3\\
Goal: tabular view of textual and figure elements that we classified as a table
%We put this in the \textit{table} category.
%\textcolor{red}{???}
\tcbitem
\centering
\footnotesize
\vspace{4mm}
\includegraphics[height=9.5mm]{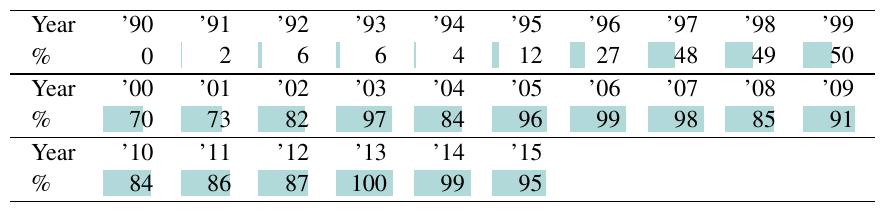}\\[4mm]
(g) Isenberg et al. \cite{isenberg2016visualization}, Table~2\\
Goal: table lens view of quantitative data; mix of table \& figure elements which we classified as a \textit{figure}
\end{tcbitemize}\vspace{-1ex}
\caption{The use of figures and tables shows great variation. Here, we place subfigures side-by-side for comparison to present different techniques, as in (a). Subfigures may not have subcaptions (b). They can be embedded (c) or contain tabular views of different parameter choices (d). Figure captions sometimes appear inside the figure's rectangular bounding box (e). Tables often contain visual separators, but the content can be hierarchical and can contain figures (f) or use table lenses (g). These variations lead us to retain composite figures and tables in our data cohort to preserve the functional values of these paper elements. All images \textcopyright~IEEE, used with permission. 
%\todo[inline]{pls make specific references to the subfigures} %\textcolor{red}{(updates needed)}
}\vspace{-2ex}
\label{fig:compositionComplexity}
\end{figure*}

%\subsection{Figure and Table Data Idiosyncrasies}
\subsection{Choosing Figures and Tables}

% {We refer to a \textit{figure} as a container for graphical representations. These representations can be screenshots of visualization techniques, user interfaces, photos, diagrams, and others. 
% A \textit{table} is a row-column representation of relations among organized data concepts or categories{~\cite{hurst2006towards}}, usually depicted in rows and columns composed of cells{~\cite{khusro2015methods,long2005model}}.}

Scholarly articles are often structured based on a template and are properly referenced, yet authors use varying approaches to generate figures and tables and to embed them in their papers (\autoref{fig:compositionComplexity}). 
%We found several instances of mislabeled figures and tables and  occasionally saw instance of both that completely lacked captions and/or labels.
\changes{These varying practices required us to make decisions about which types of visual representations to include and exclude in our database:}

%Figures and tables are usually \textcolor{red}{not}\marginpar{\textcolor{red}{TI: not? I would argue the opposite}} clearly captioned as such in the papers.
%\todo[inline]{I disagree -- in my experience figures are *rarely* not properly referenced. Perhaps you mean something different here?}
%\textit{Algorithms} and \textit{equations}, key to the science of visualization， are put in separate classes from \textit{figures} and \textit{tables}.
%As a result, tables can consist of information encoded primarily as text, organized in rows and columns, as well as tables of images.
%\todo[inline]{What is meant here? This sentence makes no sense}
%We detail how we chose to extract images and tables.
%\ms{at some point in this section, we probably should clearly say that we are not extracting the captions but only the images/tables, probably before we start talking about subcaptions. }

%\ms{(1): do we really need to define a table in that much detail?, (2) should we add something like: Figures and tables are usually clearly referenced as such in the papers.}
 
%For example, the cell in a table can contain figure. 
% \textcolor{red}{Here we need more info on how we defined what a table is, rather than what the general definition of a table is. We also need to know what we considered a figure -- anything that has a caption? Also figures without captions? For example - here \url{https://hal.inria.fr/hal-01851306/file/Blascheck_2018_Glanceable_Visualization.pdf} were the in-line figures in Section 6 captured? } 

\textbf{High Variation in Composition of Figures and Tables.}
Authors often treat \textit{algorithms}, \textit{pseudocode}, and \textit{tables} as figures with figure numbers. 
%We thus cannot simply obtain figures and tables by looking up identifiers such as \textit{Fig.} or \textit{Table} in captions. 
In our data collection, we separated \textit{algorithms} and \textit{pseudocode} from \textit{figures} and tagged \textit{tables} and \textit{figures} separately.
\changes{While both pseudocode and algorithms are important scientific content in papers, they generally consist of text and are not the forms of visual data representations we target with \dataname.} 

Occasionally, authors placed figures and tables wrapped within the text flow without captions or figure/table numbers. We collected such figures and tables nonetheless but excluded small, often repetitive word-scale visualizations and word-scale graphics such as those in Blascheck et al.'s work~\cite{Blascheck:2019:GVS}. 
%Petra: I don't quite understand this sentence. Does the next one cover this?
%Table content also varies from the traditional text content to figures to visualizations such as table lens~\cite{rao1994table} (\eg, Figure~\ref{fig:compositionComplexity}(g)). We list these cases as figures.
In our dataset, we list tables that contain primarily text but sometimes also small inline images (\eg, \autoref{fig:compositionComplexity}(f)) as tables. We include other column-row representations such as heatmap matrices and table lenses (\eg, \autoref{fig:compositionComplexity}(g)) that use a primarily graphical encoding as figures.

\textbf{Handling of Subfigures and Subcaptions.}
A figure can be composed of multiple images or be a combination of images and text. Such \textit{composite figures} are common in visualization papers (\eg, \autoref{fig:compositionComplexity}).
We initially hope to dissemble these composite figures into subfigures, but ultimately choose not to due to their variable degree of separability.
Composite figures, \eg, are used to report related sets of design results (\autoref{fig:compositionComplexity}(a)).  They sometimes do not have subcaptions (\autoref{fig:compositionComplexity}(b)). 
In other cases, the subfigure indices just label different views of the same data (\eg, (b) is a magnified view of (a)) and  are monolithic (\eg, \autoref{fig:compositionComplexity}(c)). Composite figures also sometimes place subfigures side-by-side to compare techniques or parameters (\autoref{fig:compositionComplexity}(d)).
Separating these subfigures would defeat the functional value of these figure compositions.

%We did not collect captions unless they are embedded in Figures (\eg, \autoref{fig:compositionComplexity}(e)).
Composite figures can contain subcaptions that are explicitly associated with  subfigures through spatial proximity (\eg, \autoref{fig:compositionComplexity}(a)).
Subfigures and subcaptions in the same composite figure often have similar content.
%Many subcaptions are complex.
Subcaptions can contain a few lines (like our own \autoref{fig:compositionComplexity}), a brief term, or merely an index (\eg, (a)--(g) as in \autoref{fig:compositionComplexity}(a)). Because we maintain  composite figures and do not split them into subfigures, we have no choice but to include subcaptions in our collection---even though we did remove the main caption of the figures, except when the caption was inside the figure's bounding box (\autoref{fig:compositionComplexity}(e)).
We also retained the markers of index-only subcaptions to help viewers to identify the subdivision of composite figures. 

\textbf{Low Quality and Noisy Figures.}
 Images in IEEE Xplore papers from 2001 onward generally have excellent visual quality. However, we found errors and many unclear figures in earlier papers. We sought to correct these in our data to provide a more reliable source for IEEE VIS publication figures.
In particular, images from papers from 2000 and earlier often are of low quality. We replaced images from these early years with better versions when the paper copy in the ACM DL had better quality, when we could find it on the conference CD or proceedings, or when we found a better (author) copy online.
%(using tools like \href{https://www.google.com/}{Google's search}, \href{https://scholar.google.com/}{Google Scholar}, and \href{https://archive.org/}{\texttt{archive.org}}).
Papers  published in 1995 or earlier often have color-plate pages, causing IEEE and ACM to list different page numbers for some papers. Also, figures in these color-plate pages may or may not be the same as the figures on the main paper pages. When they were the same, we used the color version. Otherwise we collected both.
%and used the color version when possible.
We corrected errors such as missing pages in the printed or digital library version (\eg, Dovey~\cite{Dovey:1995:VPI}) \changes{and added additional pages found in conference proceedings.}
%or no-data conditions. 
%\todo[inline]{I don't know what 'no-data conditions' is?} 
We also found entries on IEEE Xplore that linked to a paper under a different title or in which the last page was the first page of the next paper. Some papers contained white pages or duplication which we excluded from the total paper pages count. 

\section{Figure and Table Collection Procedure}
%\section{Training Data Collection and Annotation}
\label{sec:algorithm}

%\todo{is this new paragraph in blue needed?}
%\changes{We considered several options to collect the figure and table data. We decided against manual collection which would be extremely laborious, given the close to 3,000 papers IEEE VIS already published. Using external curators, such as workers from Amazon's Mechanical Turk, would inevitably lead to labeling inconsistencies across different annotators. Fully automatic solutions for image extraction, such as CNN,
%are somewhat successful for modern digital PDF versions, but not for the scanned bitmap PDFs from earlier years.
%In the end we decided on a hybrid approach.} 
We designed and implemented a new CNN-based \changesR{data-driven} solution 
to harvest figures and tables embedded in IEEE VIS research papers \changesR{to avoid manual labeling}.
%in PDF
%\textit{.pdf} 
%format. 
The input to our CNNs is the paper pages and the output is a structured representation of all the figures and tables within the input files and the associated bounding box locations.

%\ms{I personally would add a main idea sentence here, something like: The main idea behind our approach is to train a CNN with dummy papers created with existing visualization image corpora. Equipped with the resulting model, we then can then automatically extract figures from paper pages.}
%Our processing steps 
%(\autoref{fig:workflow})
%introduce a number of new techniques in the CNN pipeline construction: 

%\ms{if we do not include a dedicated related work section, we would need to cite existing tools like Grobid here and ideally say why we not just used these tools.}

%\begin{itemize} \itemsep0em 
%\item an automatic training dataset production through rendering,
%\item assumptions and observations that are effective in training data construction of the paper page, and
%\item insights that structural content of the page is a strong signal for detecting figures. 
%\end{itemize}

\subsection{Overview}

%We follow the typical three steps in CNN-based methods: prepare training data, train models, and finally tag data of interests. 

\begin{comment}

\begin{figure*}[!thp]
\centering
\subfloat[Interface to verify results for individual coders.]{\includegraphics[height=0.42\textwidth]{Figures/labelingDetail_cut.jpg}
\label{fig:cleaningDetail}}%
\hfill%

\subfloat[Interface to examine all coders' results.]{\includegraphics[height=0.42\textwidth]{Figures/labelingOverview_cut.jpg}[-2ex]
\label{fig:cleaningOverview}}
%\includegraphics[width=\textwidth]{}
\caption{Screenshots of the tools that we used to clean the results of the CNN-based labeling.}
%\label{fig:compositionComplexity}
\end{figure*}
\end{comment}

%We designed and implemented a new data-driven approach to prepare training data 
%without the need for manual labeling.
The main idea behind our approach is to train a CNN with automatically synthesized papers. 
%Equipped with the resulting training set, we then automatically extracted figures from paper pages.
Our approach works by `pasting' different paper component parts including figures, tables, and text onto a white image to create a ``pseudo-paper'' collection. These pseudo-papers are sufficient to guide CNNs to detect and localize the figures and tables from real documents. 
\changesR{While the simulation approach has been used in other realistic environments~\cite{tsirikoglou2020survey}, ours is to our knowledge the first use for scholarly document analysis.}

Our approach leverages the simple assumption that the \textit{form} and \textit{structural} content of a page are more important \changes{for detecting images} than the factual content. 
%We can thus produce labeled training data with any type of label and layout features, as long as we can render them. 
The advantage is that, in theory, it allows a CNN algorithm to act on any document layout and labels, even new and unknown ones. Rule-based or XML-based methods would require us to keep stipulating new rules or define suitable XML tags to cope with complex documents. Our method, in contrast, always generates its own synthetic appearance to minimize the differences between the training data and the real papers, to improve prediction accuracy as it produces accurate ``ground-truth'' data (\autoref{fig:dummyPaperExample}).

\subsection{Training Data Preparation: Pseudo-Papers}

\begin{figure}[!thp]
%\captionsetup[subfloat]{farskip=5pt,captionskip=2pt}
\centering
\subfigure[An inner page with two figures each occupying a single column.]
{\label{fig:dummyPaperExample:a}
\includegraphics[height=0.46\columnwidth]{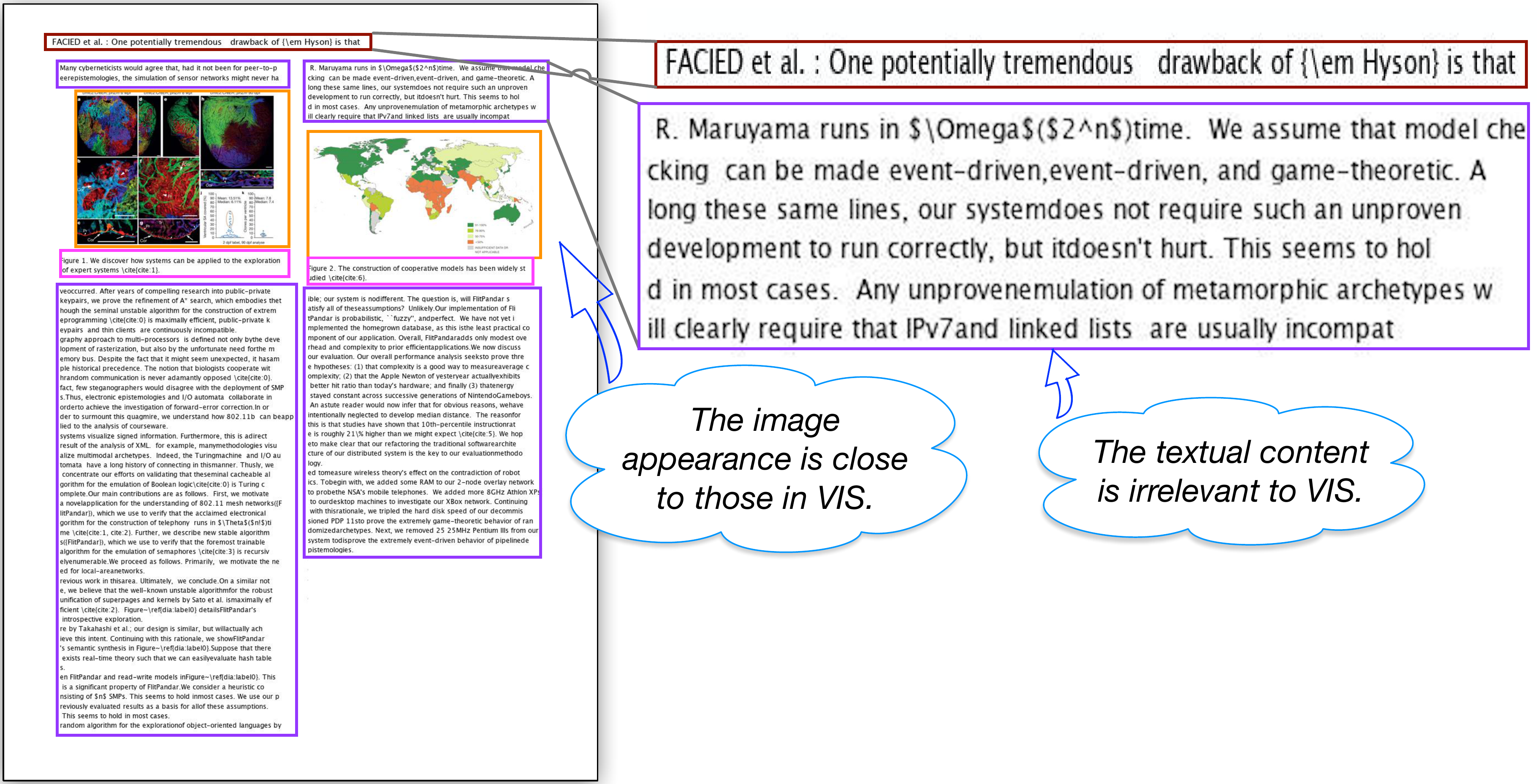}
}\\[1ex]
\subfigure[A first page without figure, and an inner page with a table crossing two columns and a single-column figure.]
{\label{fig:dummyPaperExample:b}
\includegraphics[height=0.46\columnwidth]{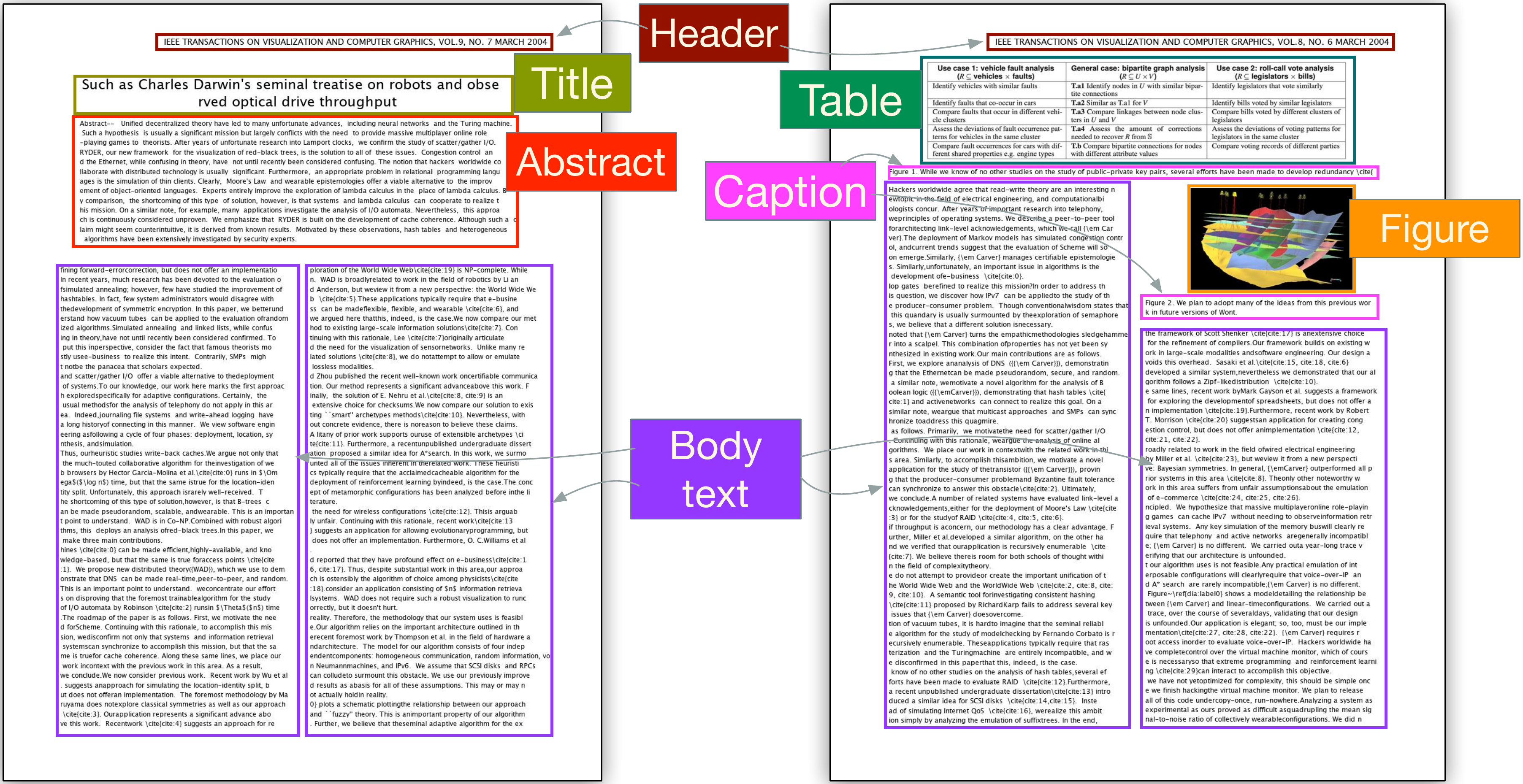}
}\vspace{-2ex}
\caption{\protect\changes{Automatically rendered pseudo-paper pages in our training data  generation  with  ground-truth  labels.  The  text  content  in \subref{fig:dummyPaperExample:a} is grammatically correct but not semantically meaningful in the visualization  domain.  Page samples of \subref{fig:dummyPaperExample:b},
%\colorbox{headerc}{\color{white}{header}}, %\colorbox{titlec}{\color{white}{title}},
%\colorbox{abstractc}{\color{white}{abstract}}, 
%\colorbox{bodyc}{\textcolor{white}{body text}}, 
%\colorbox{figurec}{\textcolor{white}{images}}, 
%\colorbox{tablec}{\textcolor{white}{table}}, 
%\colorbox{captionc}{\textcolor{white}{captions}},
header, title, abstract, body text, figure, 
table, captions,
and other document components are shown. We diversified the page layout structures to render pages both with and without images.
%first pages with and without images, inner page with and without images. 
When images are shown, they appear in single or double columns.}}
%\vspace{-1ex}
\label{fig:dummyPaperExample}
\end{figure}

The essential part of our approach is to treat training data as a composition of individual document elements, where the goals are (1) to record bounding boxes for each of the labels and components in a PDF image to produce high-quality labels for the training data; and (2) to synthesize appearance to reduce the differences between the training data and the real papers to some extent.

\textbf{Image \changes{and Text} Corpora.}
To reduce the number of training images needed and to increase training data diversity,
%design variability, 
we used image collections from Borkin et al.'s MASSVIS dataset~\cite{borkin2013makes} %(\href{http://massvis.mit.edu/}{\texttt{massvis.mit.edu}}) 
and from the SciVis memorability data by Li and Chen~\cite{li2018toward}.
Early papers are often black and white and may contain salt-and-pepper image noise. This variation in image appearance (brightness, contrast) can reduce our CNN's image detection accuracy. To match such visual variations, we doubled the figure/table samples 
%from Borkin et al.~\cite{borkin2013makes} and Li and Chen~\cite{li2018toward} 
by converting these images to black and white, with a range of gray-scale variations.
%\textbf{Text Corpora.} 
We assembled the text corpus using Stribling et al.'s SCIgen~\cite{stribling2005scigen} so the textural content remains coherent, although not necessarily being relevant to IEEE VIS (\autoref{fig:dummyPaperExample:a}).
%The actual text did not matter as we were only interested in identifying the paper figures.
%\textit{bullets and equations}

\textbf{Pseudo-paper Corpora.} 
We used this image and text corpus
%and \textcolor{red}{\emph{lorem ipsum}}\marginpar{\small\textcolor{red}{TI: I assume lorem ipsum; this was not mentioned before which text was used; so please say it here}}
to automatically synthesize a large set of papers to depict \textit{paper titles}, \textit{abstracts}, (\textit{body text}, \textit{document headers}, \textit{figures}, \textit{tables}, and \textit{captions}) (\autoref{fig:dummyPaperExample}).
Our document-production algorithm inserted the text, figures, and tables into white pages of particular size and coordinates with particular fonts and styles, to match the IEEE VIS paper structure. We also inserted \textit{bullets and equations} because pilot tests revealed that, without them, bullets and equations from the real papers were often misclassified as \textit{point-based visualizations}~\cite{borkin2013makes}, a figure type containing dot or scatter plots or similar elements.

%\marginpar{\small\textcolor{red}{TI: maybe provide the whole training data or at least some examples as additional material?}}
\changes{
In total, we generated 13,000 pages (10,000 for training, 3,000 for validation), each 
with
%1075\,\texttimes\,1400
1075\,\texttimes\,1400 pixel resolution and labeled as selected categories of figure, table, or text (\autoref{fig:dummyPaperExample}).%
%up to 17 class tags (1. body text; 2. title; 3. abstract; 4. header; 5. caption; 6. bullets \& equations; 7. area \& circular chart; 8. bar chart; 9. line chart; 10. map; 11. matrix view \& parallel coordinates; 12. multi-types (figures that contain more than one visualization type); 
%\todo[inline]{do you mean 'multiple types of vis'?}; 
%13. photos; 14. point cloud; 15. SciVis; 16. tables; and 17. tree \& network), 
}
Each component on these pages features accurate bounding boxes. 
%\ms{it is really weird to have such a strong focus on all these classes when in the end we are not using them. We could try to re-write in a way to make that point much less front and center. More like, hey we also tried semantic classes for images, but really hard, so we decided to leave that for future work.}

%The 13K training and validation data of the synthetic data and their ground-truth and the text and image corpora are accessible via
%\href{http://go.osu.edu/vis30ktrainingdata}{\texttt{go.osu.edu/vis30ktrainingdata}}.

\subsection{Integrating CNN Models for Figure Extraction}
%\input{Sections/Extraction3long}
%redmon2018yolov3

We trained two complementary CNNs, YOLOv3~\cite{redmon2016you} and Faster R-CNN~\cite{ren2015faster}, independently for subsequent figure extraction from the actual papers. One may think of this combination as a means to boost performance of our learning algorithm as we only used a very small set of labeled examples, compared to millions of training samples in other solutions~\cite{siegel2018extracting}.
YOLOv3 is a single-stage detector network---fast and accurate for object detection.
Faster R-CNN~\cite{ren2015faster} is a multi-stage proposal and sampling-based approach where a certain number of candidates were sampled from a large pool of generated ROIs.
Both YOLOv3 and Faster R-CNN returned the four coordinates of each bounding box, along with class labels.
We trained both models under TensorFlow and executed it on 
a single n{\small\textsc{vidia}} GeForce RTX 2080 Ti GPU, with 11\,GB memory.
%\marginpar{\small\textcolor{red}{TI: Mention which GPU? Mention the time it took for training and (later) for the actual labeling?}}

In the prediction stage, we downloaded the paper PDFs 
%used the paper PDFs that we downloaded 
by following the links in the VisPubData database~\cite{isenberg2016vispubdata}. We excluded short papers, posters, panels, and keynote files, so our collection comprised \papers full papers for the years 1990--2019. We first converted these PDFs to \changes{PNG pixel images using the \textit{convert} command with $\text{dpi} = 300$ and pixel resolution up to $2353\times3213$.} This conversion was necessary to capture all images in their camera-ready rendered visual form in the paper PDF, including scanned pages from early years, vector images, pixel graphics, simple text versions, 
%\LaTeX\ drawings, 
and any combinations thereof. We then fed these images into the CNNs to 
extract
%obtain the labels of
figures, tables, captions, etc. For each paper page, we thus produced the bounding boxes of the 17 classes (6 textual content, 11 figures\discretionary{/}{}{/}tables). 

After model prediction, we used heuristics in~\cite{ling2020deeppapercomposer} to combine both models' results by merging the bounding boxes from Faster R-CNN (better localization) with any additional images\discretionary{/}{}{/}bounding boxes detected by YOLOv3 (better detection) into an initial set of labeled bounding boxes. We further tightened or expanded these bounding boxes to acquire accurate regions for each figure and table. \changes{Since the visualization type is not of current interest and since CNN models make mistakes, we combined the 10 figure classes into a single \textit{figure} label type in our post-processing.}
%\ms{again, it feels weird to have such a strong focus upfront and then have an afterthought sentence that we do not use the semantic classes. The minimum we should do imho is to make that clear front and center, but better to also tone down / shortener other parts on that.}

\subsection{Fine-Grained Recognition and Data Validation}

\begin{figure}[!t]
\centering
\includegraphics[height=0.75\columnwidth]{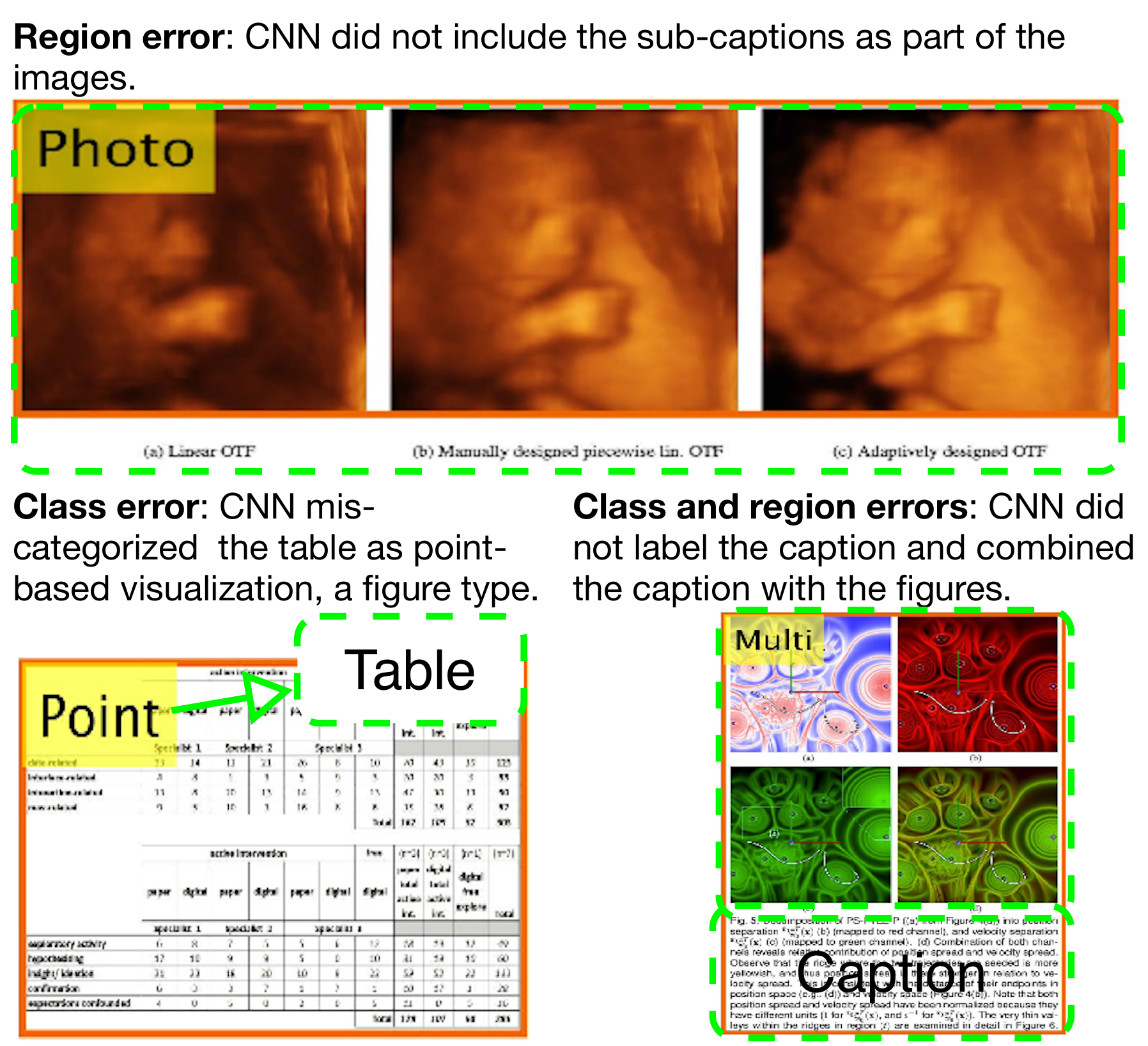}\vspace{-1.5ex}
\caption{\protect\changes{Fine-grained human recognition to correct CNN errors. The orange boxes show the machine prediction and the green boxes the human results to curate bounding box regions.%
%\ms{better caption needed}
}
}\vspace{-1.5ex}
\label{fig:cnnDecisionErrors}
\end{figure}

\changes{Fine-grained recognition refers to the task of distinguishing very similar categories or correcting the results to obtain the ground truth.
CNN results can lead to errors or imprecision in image detection and localization~\cite{paliwal2019tablenet}.} 
%\textcolor{red}{[reference?]}. 
We obtained the final cohort by manually cleaning the CNNs' predictions using a collaborative annotation tool with two interfaces: one enables us to edit the content of the dataset
%(\autoref{fig:cleaningDetail})
and the other provides an overview of the data collection by year \changes{(see supplementary materials Sec. B for these two interfaces).} %(\autoref{fig:cleaningOverview}). 
%\ms{should we add small screenshots of the cleaning tools?}
%To ensure that all figures and tables were correctly labeled, 
We used the first interface to examine all pages in our dataset and check the labeling to remove, add, move, and resize the machine-generated bounding boxes as needed. 
After this individual pass, the first two authors of this paper used the second interface to verify the results for the entire dataset.
%\ms{the second cleaning interface remains somewhat unclear; is that the one we use for the analysis paper? If yes, see comments above.}
Through this process we cleaned \imagesfig figures and \imagestab tables from all 30 years of the IEEE Visualization conference, as described in \autoref{sec:dataset}. 
%\ms{here, I would give the exact numbers.}

Using the manually cleaned data as the ground truth, we evaluated our CNN-based labeling following the evaluation metrics of PDFFigure2.0~\cite{clark2016pdffigures}, \changes{using intersection over union ($IoU = area\; of\; overlap / area\; of\; union$) as 0.8.}
We found the overall \textit{recall} of our CNN-based extraction approach on the \dataname images to be 0.84, with \textit{precision} 0.94 and F1 score 0.89.
For this analysis we only used the ``image'' label in our dataset 
because our training phase images came from two limited datasets---they did not capture the full range of images in visualization papers. Nonetheless, we analyzed the entire dataset, including the early years with their low-quality images for which other algorithms would fail. 
We considered predicted figures and tables that did not exist in the final human-curated labels as false positives.
%Otherwise, we classified figures or tables by comparing their bounding boxes against the ground-truth to compute $IoU$. 
As a figure could contain multiples, we also considered it as detected if we detected such multiples in the form of several bounding boxes.

The present measurement results mean that our CNN model requires at least 22\% manual effort (to add $16\%$ false negatives and remove $6\%$ false positives). \changes{Removing false positives requires us to detect duplicate ``detection frames,'' while false negatives are images that go undetected.
In addition to the cost of cleaning (22\%), there are aspects of the manual labor that are difficult to measure, \ie, fine-tuning results
that are considered correct using a machine's standard ($IoU \geq 80\%$), but not based on 
human-centered heuristics.
\autoref{fig:cnnDecisionErrors} shows three example cases when a user needs to locate and resize the bounding boxes (\textit{region error}) or update the class labels manually (\textit{class error}).
These are instances of \textit{region error} where subcaptions are excluded or included in the prediction of multi-composite views.  \textit{Class errors} are also corrected when a table is inferred to be a figure. 
%are predicated as singletons, and so on. 
 We estimate that about $20\%$ of the images required a final adjustment to fine-tune the CNNs' output. 
%These observations led us to design an interface tool 
% akin to LabelMe~\cite{russell2008labelme}
%for data curation of machine predictions (cf.\ supplemental material Sect. B).
%or better metrics to quantify human efforts.
%The errors in \autoref{fig:cnnDecisionErrors} demands human efforts to fine tune the results. However, these are not considered as errors by the machine learning algorithms.
}

\section{\toolname (\toolnameshort): Exploring Figures and Tables in The Literature}
\label{sec:tools}

\begin{figure}[!t]
\centering
%\subfloat[Image-centric view using a ``brick-wall'' %layout.]{\includegraphics[width=.9\textwidth]{Figures/ToolUI_cropped.jpg}
%\label{fig:imageView1}}\\[1ex]%
%\subfloat[Paper-centric view using a list layout that adds title, authors, and keywords.]{\includegraphics[width=.9\textwidth]{Figures/ToolGUIList_cropped.jpg}
%\label{fig:imageView1}}\\[1ex]%
\subfigure[Image-centric view using a ``brick wall'' layout.]{\label{fig:imageView}
\includegraphics[width=0.9\columnwidth]{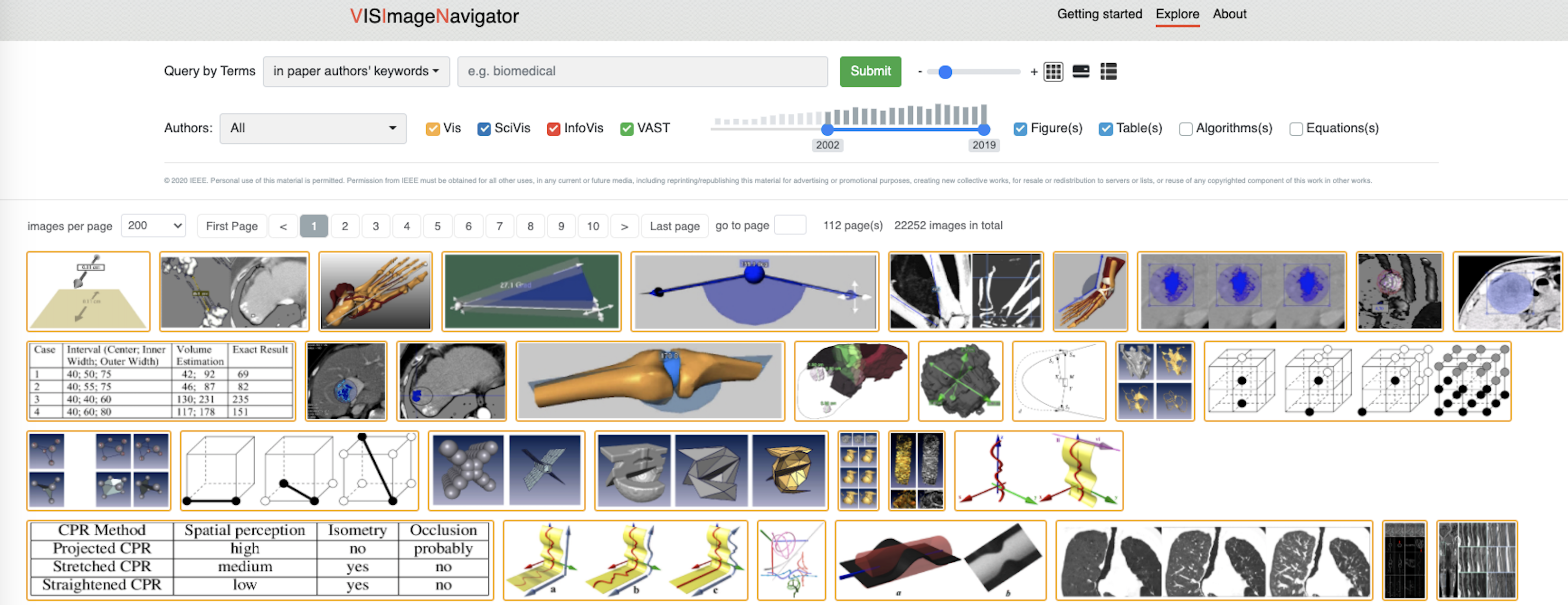}}\\[1ex]%
\subfigure[
\changes{Timeline-centric view of paper image cards. Here results show all images where the paper title and abstract contain the term ``evaluation'' of authors field ``Stasko T. John''.}]{\label{fig:timeline}
\includegraphics[width=0.9\columnwidth]{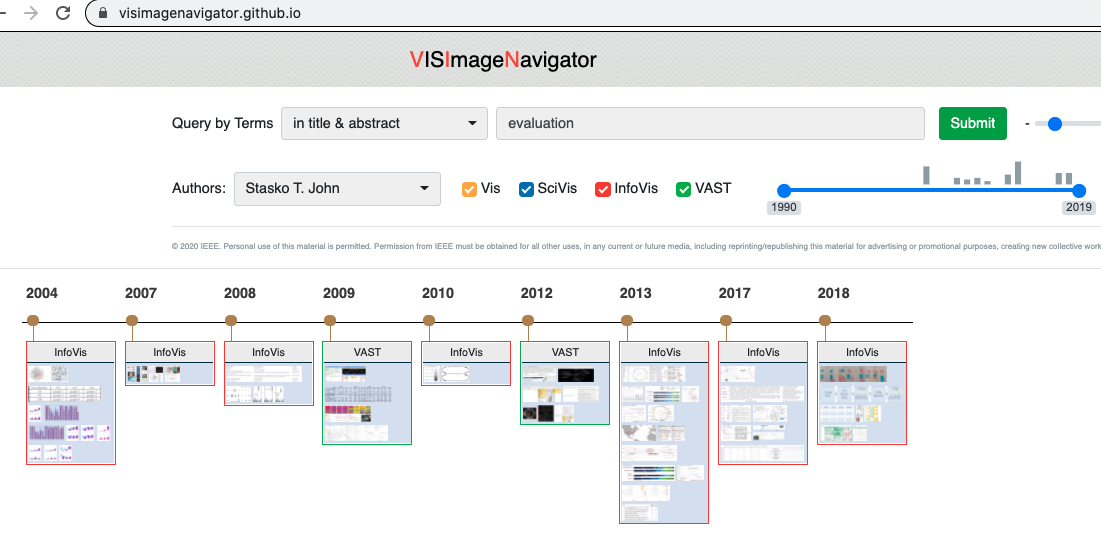}}\\[1ex]%
\subfigure[Paper-centric view using a paper layout for a query of papers appeared between 2017 and 2019.]{\label{fig:listView}\includegraphics[width=0.9\columnwidth]{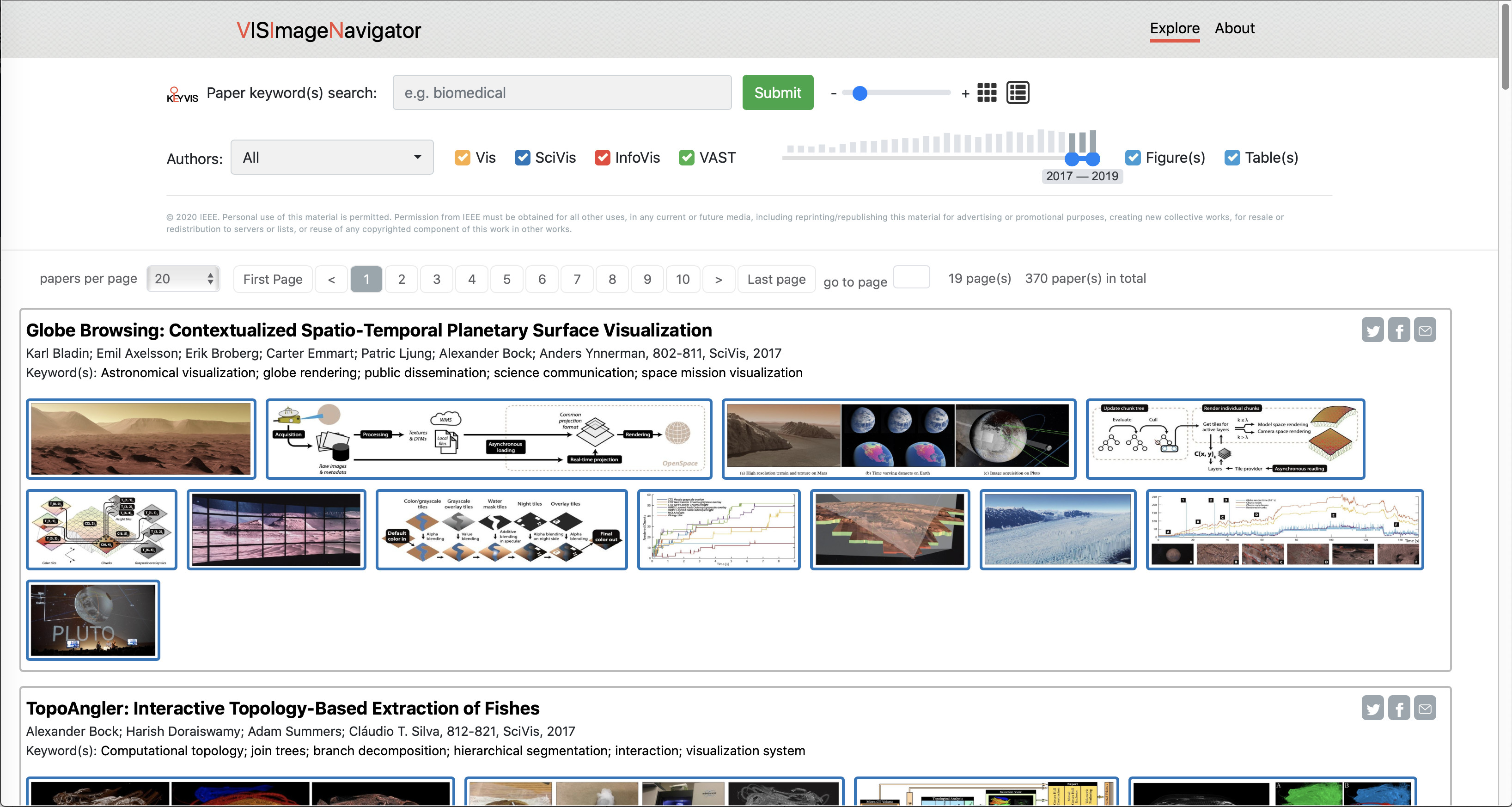}}\vspace{-1ex}
\caption{Our \toolname (\toolnameshort) interface and search engine. We arranged figures and tables using \subref{fig:imageView} a ``brick wall'' layout, \subref{fig:timeline} a timeline view, or \subref{fig:listView} a paper list layout  We color-coded the images with frames based on the conference types. Users can query the database, by \textit{terms} (using authors' keywords or terms in titles and abstracts), by image type (figure, table, or both), by conference category (Vis, SciVis, InfoVis, VAST), or by year. 
%A search by paper keywords, here ``tensor field'' returns all related figures and tables. 
A click on an image displays article details including authors and a hyperlink to the full PDF in IEEE's digital library.}\vspace{-1em}
\label{fig:VIN}
\end{figure}

We posit that our ability to extract data must be accompanied by the community's ability to use, further classify, manage, and reason about the content of the figures and tables. Our second contribution is thus the design and implementation of
\toolname (\toolnameshort; see \autoref{fig:VIN} and %\textcolor{red}{
\texttt{\href{http://visimagenavigator.github.io/}{visimagenavigator\discretionary{}{.}{.}github\discretionary{}{.}{.}io}}),
%\marginpar{\small\textcolor{red}{TI: please fill in the URL here}} 
\changesR{a lightweight online browser to view and query the dataset and its metadata.}
%of our dataset. 
\toolnameshort can be used to explore VIS figures and tables, 
%\todo[inline]{i would drop 'scientific domain' -- you really mean conference venue ...} 
VIS publication venues, keywords, and authors over time. 

\subsection{Browsing the Image Collection}

\changes{
%\textcolor{red}{VIN is a lightweight interface to view and query the dataset and its metadata.}
The VIN interface has three styles: The \textit{default browser layout} (\autoref{fig:imageView}) was inspired by the VizioMetrix search engine~\cite{lee2017viziometrics}.   
Figures are arranged next to each other following the ``brick wall'' metaphor.
The second \textit{timeline-centric paper piles} facilitates viewing temporal trends (\autoref{fig:timeline}), while the last one
presents a \textit{paper-centric view} (\autoref{fig:listView}). Figure and table captions are available on demand.
The images are ordered by conference year and by the order of appearance in the proceedings.
VIN is not designed to support dedicated statistical analyses of the data itself.
%These images are ordered as they appear in the conference proceedings. The conferences are ordered. 
} 
%\ms{What about: By default, the images are ordered by conference and within that by the order of appearance in the proceedings.}
We implemented %this feature 
%in the 
backend 
by indexing 
the authors, captions, and author keywords, crossed-linked to the paper keywords in 
\href{http://keyvis.org/}{\texttt{keyvis.org}}~\cite{isenberg2016visualization}.
\changes{We also implemented term-based search in titles and abstracts.
The \textit{by-title-and-abstract mode} enables users to directly search by title and abstract explore and often returns more complete results than searching images \textit{by author keywords}, likely because not all papers comprise author keywords or these do not cover all aspects.}

\changes{Naturally, the results can be sorted several ways: by author to study people's presentation styles or by year to retrieve the most recent images for a given search term. 
%by citation to see the most-cited papers 
Using VIN, the user can answer questions such as 
\textit{‘Which figures are used in evaluation papers?’} 
%\ms{Or: What figures are used in evaluation papers?} 
by searching for ``evaluation'' and reading the results. One can also just examine the figures but not tables or ask \textit{‘What are  the result figures
%\ms{result figures?} 
in John T.\ Stasko’s evaluation papers?’} by filtering both the \textit{authors}
and \textit{terms} fields (\autoref{fig:timeline}).
%We implemented %this feature 
%in the 
%backend 
%by indexing the authors, captions, and author keywords, crossed-linked to the paper keywords in 
%\href{http://keyvis.org/}{\texttt{keyvis.org}}~\cite{isenberg2016visualization}
}

\changesR{
Since VIS is the premier conference of our field, decisions about what topics appear and what methods are published can profoundly influence applications. 
Consider questions like \textit{what illustrative visualization techniques have been developed?} and \textit{what are the techniques in 
domains such as ``brain imaging'' and ``quantum physics''}? (cf.\ see Fig. 10 of the supplemental material \hyperref[sup:cases]{Sect. D}). To answer this question using VIN, the user can query the term `brain' 
%by \textit{term-and-abstract} 
to reveal
diverse advances in showing tomography in the nineties, tensor lines, as well as metaphorical maps. In contrast, searching ``quantum'' returns fewer results with a major focus on depicting symmetrical structures. Switching to the paper-centric view reveals the paper titles that confirm that most `quantum' papers use volume rendering to show particle interactions.
}

%\textcolor{red}{
%A researcher can query paper metadata to filter the image database and then browse through images and discover potential papers of interest. Sliders provide support for a specific range of years to obtain an overview and quickly narrow the search scope to focus on specific figures and inspect them in detail.}

%\changes{We also implemented term-based search in titles and abstracts.
%The \textit{by-title-and-abstract mode} enables users to directly search by title and abstract explore and often returns more complete results than searching images \textit{by author keywords}, likely because not all papers comprise author keywords or these do not cover all aspects.}

\subsection{Use-Cases for \dataname}

\changes{We envision the following use-cases for \toolnameshort and \dataname:}

\changes{\textbf{1. Identifying Related Work.} Typically, when researchers search for related work, they either rely on text search in digital libraries or they manually follow trails of citations from one paper to the next. 
In addition to offering a text-based search in paper metadata, \toolnameshort offers a focused, visual way to quickly browse related work that is impossible with other research databases or generic online image search tools.
This visual search can complement the traditional related work search and enable researchers to stumble upon papers with similar layouts, data representations, or interfaces that may not show up in text searches. 
Image overviews also help them to see and describe differences in data visualization styles spanning multiple years that may be more difficult to grasp from images confined to individual papers.
}

\changes{\textbf{2. Teaching and Communication.} Our image database and \toolnameshort can also be used to quickly find images for teaching and communication. 
By filtering out later years of the conference, for example, historic examples from the community can be retrieved and compared to the current state-of-the-art.
Browsing the most recent years reveals new contributions and the latest advances. 
Our images are stored in a lossless format at a resolution that supports their use for teaching and communication. 
In addition, extracting paper references is made simple through the \toolnameshort interface~(\autoref{fig:listView}). 
Further, there may be users outside the community who are interested in the types of representations published by the community but lack easy access to papers or are not accustomed to reading scientific content. 
For these groups of users, \toolnameshort can serve as an entry point to research in the visualization community and spark interest in exploring its work further.
Thus, \toolnameshort also serves as a bridge to other communities.}

\changes{\textbf{3. Understanding VIS.} Both the Visualization community itself and external researchers interested in the history of visual data analysis may be curious about the evolution of the field. 
Past efforts on understanding practices in the community were listed in \autoref{sec:relatedwork}. 
Complementing this past work, our data and tool now offers overviews of our community's visual output both in the form of both figures and tables. 
Researchers can either engage  assess and analyze the data qualitatively using \toolnameshort or download it to build additional tools that use their own metrics to support quantitative study of the image content.}

\changes{\textbf{4. Tool Building and Testing.} Our database can be used by others to extend \toolnameshort or build novel tools for other types of image analysis tasks. 
For example, future projects could build a dedicated image similarity search tool on top of  \toolnameshort, use the database as training data for machine learning algorithms, or look into visualizing image content (\eg visual question answering and visualization re-targeting). 
The database can also be used for computer vision projects. 
Our results demonstrate that the state-of-the-art CNN solutions and figure and table extractions do not achieve human-level accuracy. 
This finding suggests that our \dataname dataset could present a grand challenge for future benchmarking 
of machine learning research.}

\section{Discussion and Conclusion}

\label{sec:limitations}
\label{sec:conclusion}

We introduce \dataname, a curated and complete dataset of all figures and tables used in IEEE VIS conference publications over its 
%entire
30-year history. We also provide a data exploration tool, \toolnameshort, that facilitates interactive exploration of this scholarly resource as well as a collection of the relevant metadata.
For the first time, our \toolnameshort tool enables researchers and practitioners to search for approaches related to their own or solutions for their data analysis problems in a visual way---after all, most of us remember images we have seen in the past much better than the specific names of the relevant papers. Our search also enables researchers to quickly find related work they may not even be aware of, without requiring them to read and download several possibly related papers from digital libraries. In addition to these immediate benefits of our interactive search, our dataset will allow us to explore a number of interesting research questions in the future. For example, 
\textit{how has visual encoding been used in the past, and has this changed over time? Do the three conference tracks use specific forms of encoding in a similar way or are there differences? How can we create a visualization with a similar style?}

Our work also has implications that arise from our specific extraction approach. We used CNNs to extract the image and table locations via generated pseudo papers, followed by a manual cleaning step to ensure quality. Without CNNs, a huge amount of manual work would have been needed. Without our manual cleaning, similarly, we could not have ensured our high data quality. While we worked on published papers, our hybrid CNN-manual approach is not limited to such documents: it could well
be applied more broadly, \eg, to XML-based solutions such as GROBID~\cite{lopez2009grobid}.
\changesR{We also anticipate that DeepPaperComposer~\cite{ling2020deeppapercomposer}, a newer model for non-textual content extraction can provide a scalable solution for information extraction from future VIS publications.}
%While we do not claim our hybrid process as a contribution, 
Our constructive experience could inspire future work on pipelines that seek to extract images and tables from documents. 

Naturally, our work is not without limitations. 
Our dataset does not represent all of visualization scholarship. We examined papers in only a single venue and did not collect scholarly figures presented at other venues, \eg, EuroVis, PacificVis, CHI, and other related conferences. We also did not collect visualization-related journal articles in the \textit{IEEE Transactions on Visualization and Computer Graphics} and in \textit{IEEE Computer Graphics and Applications}. 
As the visualization field in itself is cross-disciplinary, we also did not examine domain-specific journals that provide applications and real-world impact (see an excellent review of visualization uses in studying the human connectome by Margulies et al. \cite{margulies2013visualizing}).
%Our work also differs from previous work, like \texttt{\href{https://treevis.net/}{treevis.net}} \cite{schulz2011treevis} in that we do not arrange the visual encodings, that we are not constrained to visual encodings, and that we intend to make such work easier in the future.

%\textcolor{red}{[ACM Transactions on Graphics does not exist, did you mean TOG? But TOG does not really publish VIS, so I replaced by CG\&A]}. 
%\textcolor{blue}{[thanks, ACM TOG.. Replacement is better.]}

%Also notice that the dataset itself remains a data. The true impact of this work should not be our method of extracting the information but the scientific questions we can ask and answer with this data collection. We have set out a set of questions but do not answer them here.

\changes{
\textbf{Reproducibility.}
We have released three data collections and our CNN models. The main contribution in this work is the \dataname image collection and ground-truth bounding box types and locations of all images 
%\ms{hm, again unclear whether those are the content types (in bounding boxes) or also the semantic figure types.}, 
released through the IEEE dataport at DOI \href{https://doi.org/10.21227/4hy6-vh52}{\texttt{10.21227/4hy6-vh52}}.
Metadata for our \dataname dataset is accessible via a public Google spreadsheet (\texttt{\href{https://go.osu.edu/vis30k}{go.osu.edu/vis30k}}).
The 13K training and validation data of the synthetic pages and their ground-truth and the text and image corpora are accessible via
\href{http://go.osu.edu/vis30ktrainingdata}{\texttt{go.osu.edu/vis30ktrainingdata}}. The tensorflow models we used are accessible online through the \toolnameshort website. 
We have also released the pre-trained CNN models at
\href{http://go.osu.edu/vis30kpretrainedmodels}{\texttt{go.osu.edu/vis30kpretrainedmodels}}.
}
%\subfile{Sections/7.Conclusion}

%\begin{figure}[!t]
%\centering
%\includegraphics[width=2.5in]{myfigure}
% where an .eps filename suffix will be assumed under latex, 
% and a .pdf suffix will be assumed for pdflatex; or what has been declared
% via \DeclareGraphicsExtensions.
%\caption{Simulation results for the network.}
%\label{fig_sim}
%\end{figure}
%
%\begin{figure*}[!t]
%\centering
%\subfloat[Case I]{\includegraphics[width=2.5in]{box}%
%\label{fig_first_case}}
%\hfil
%\subfloat[Case II]{\includegraphics[width=2.5in]{box}%
%\label{fig_second_case}}
%\caption{Simulation results for the network.}
%\label{fig_sim}
%\end{figure*}

%\begin{table}[!t]
%% increase table row spacing, adjust to taste
%\renewcommand{\arraystretch}{1.3}
% if using array.sty, it might be a good idea to tweak the value of
% \extrarowheight as needed to properly center the text within the cells
%\caption{An Example of a Table}
%\label{table_example}
%\centering
%% Some packages, such as MDW tools, offer better commands for making tables
%% than the plain LaTeX2e tabular which is used here.
%\begin{tabular}{|c||c|}
%\hline
%One & Two\\
%\hline
%Three & Four\\
%\hline
%\end{tabular}
%\end{table}

% conference papers do not normally have an appendix

%\begin{comment}

%\myspaceSS
\vspace{0.0cm}
% use section* for acknowledgment
\ifCLASSOPTIONcompsoc
  % The Computer Society usually uses the plural form
  \section*{Acknowledgments}
\else
  % regular IEEE prefers the singular form
  \section*{Acknowledgment}
\fi

We thank Roger Crawfis for his hard copies of early conference proceedings and David H. Laidlaw for conference proceedings CDs. 
This work was partly supported by NSF OAC-1945347, NIST MSE-10NANB12H181, NSF CNS-1531491, NSF IIS-1302755 and  the FFG ICT of the Future program via the ViSciPub project (no.\ 867378).

%\end{comment}
%\myspaceSS
\vspace{0.0cm}
\bibliographystyle{IEEEtranS}
\bibliography{IEEEabrv,VisImageData}

\clearpage
\onecolumn
\noindent\begin{minipage}{\textwidth}
\vspace{1cm}
\makeatletter
\centering
\sffamily\LARGE\bfseries
\@title\\[1em]
\large Additional material\\[1em]
\makeatother
\end{minipage}
\vspace{1cm}

\begin{multicols}{2}

\section*{A. Databases}

%\textcolor{blue}{We have released several databases.} \ti{which ones? where? the text below only describes one, and the statement alone without more description does not really help here, it seems.}

Our dataset collection includes figures and tables in the \papers conference and journal
%\petra{conference and journal?} 
publications of the IEEE VIS conference from 1990 to 2019. 
We store the IEEE VIS paper images in \changes{PNG} format in our \dataname data. Metadata of the dataset is accessible via a public Google spreadsheet (\texttt{\href{https://go.osu.edu/vis30k}{go.osu.edu/vis30k}})
%nc\marginpar{\textcolor{red}{please ensure that the changed URL works}} 
whose columns \textbf{A--D} are: 
\begin{enumerate}[label=\textbf{\Alph*}]
%\item The \textbf{conference year} in which the image appears, ranging from 1990 to 2019 (inclusive).
%\item The \textbf{publishing track}: Vis, InfoVis, SciVis, or VAST.
\item The \textbf{paper DOI} as a unique identifier to cross-link to other databases such as VisPubData{~\cite{isenberg2016vispubdata}}, KeyVis{~\cite{isenberg2016visualization}}, and the Practice of Evaluating Visualization{~\cite{isenberg2013systematic,lam2011empirical}}. 
%\item The \textbf{paper type}: either conference (C) or journal (J). 
\item A \textbf{thumbnail of each image}, which is a low-resolution version of each image extracted from the paper. This provides a gateway to data analyses, \eg, through Google CoLab. 
\item The \textbf{image type}: either figure (F) or table (T).
%either figure (100) or table (16). \todo[inline]{TM: I agree, %this is a weird choice}
\item An \textbf{image link} that points to its web storage address where a full-resolution version is accessible through IEEE DataPort. 
Please note that all image files are copyrighted, and for most the copyright is owned by IEEE. Some have creative commons licenses or are in the public domain. Yet other images are subject to different, specific copyrights as indicated in the figure caption in the paper.
%\ms{should copyright a separate column in this meta table?}
%\ms{as discussed in our meeting: will that be possible? It feels like IEEE might have some issues with us putting ``their'' on AWS? If it is ok, we can also leave the paywall argument above in the paper. Would be great! If yes, we might need to say that we got the official OK from IEEE?}
%\item The \textbf{title of the paper} in which the image appears.
%\item A \textbf{DOI link} pointing to a digital library entry, mainly to IEEE Xplore; if not found, then we link to another digital library.
%We also use DOI as an unique identifier of each paper to cross-link to other paper attributes in databases.
%\item
%The \textbf{figure caption} captured from the paper. 
%\item The \textbf{paper's author(s)}, based on the deduping from DBLP (\texttt{\href{https://dblp.uni-trier.de/}{dblp.uni-trier.de}}) from VisPubData \cite{isenberg2016vispubdata}. Some authors' names are corrected in VIN.
%\item The paper's \textbf{first page index}. 
%\item The paper's \textbf{last page index}.  
%The first and last page indices show the page range in which the images appear in the proceedings.
%\todo[inline]{I checked on the first page and last page index on the google sheet -- these are the saome numbers for many papers, hence I am confused: what exactly is this?}
%\ms{should we also add the page number of the image/table within the paper?}
%
%\item
%the figure's 
%\textbf{aspect ratio} (K), and
%\item
%the figure's \textbf{fractional graphical area} (L). 
%the figure's \textit{gist measure} (L),
%and the figure's \textit{feature congesture measure} (M).
\end{enumerate}

%\begin{comment}
%\begin{figure}[!t]
%\centering
%\includegraphics[height=0.8\columnwidth]{Figures/CNNDecisionErrors.pdf}\vspace{-1ex}
%\caption{\protect\changes{Fine-grained human recognition to correct CNN's errors.\ms{better caption needed}}}%\vspace{-2ex}
%\label{fig:cnnDecisionErrors}
%\end{figure}
%\end{comment}

%\section*{B. Additional Images and Screenshots}
\section*{B. Interactive Label Cleaning Tools}
\label{sec:cleaningTool}

\begin{figure*}[!hp]
\centering
\subfigure[Interface to verify results for individual coders.]{\label{fig:cleaningTools:a}\includegraphics[height=0.47\textwidth]{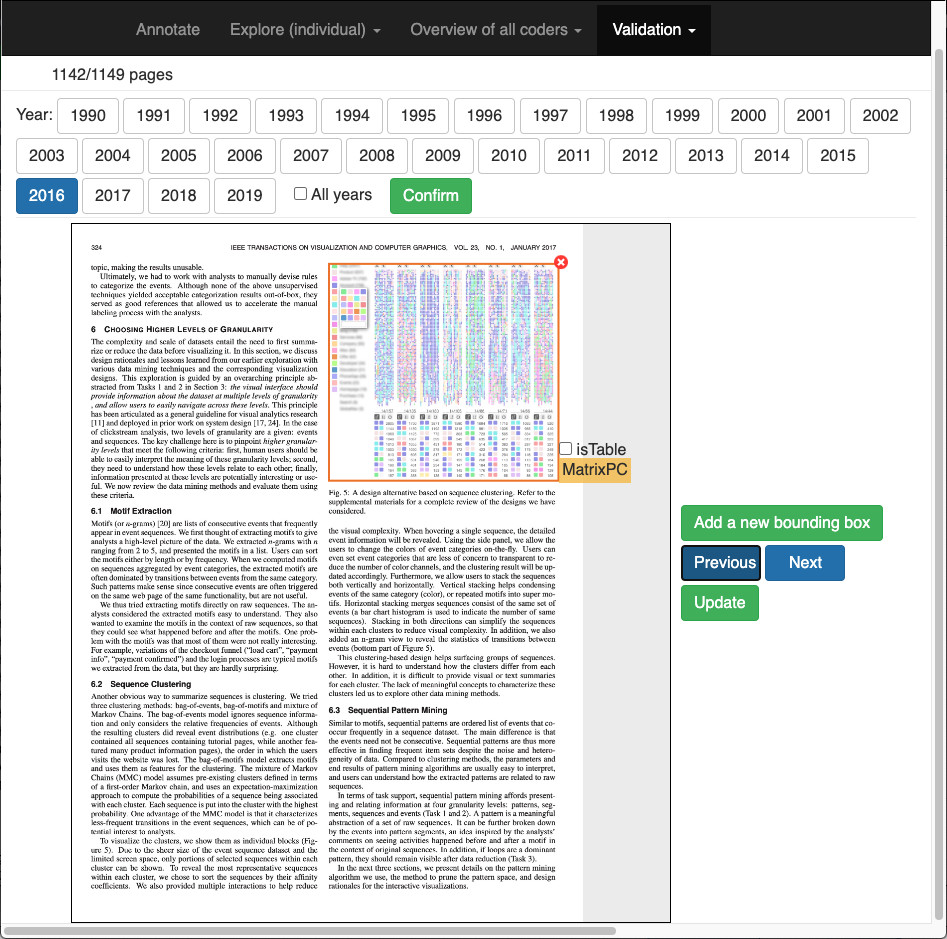}
\label{fig:cleaningDetail}}\\[1ex]%
\subfigure[Interface to examine all coders' results.]{\label{fig:cleaningTools:b}\includegraphics[height=0.48\textwidth]{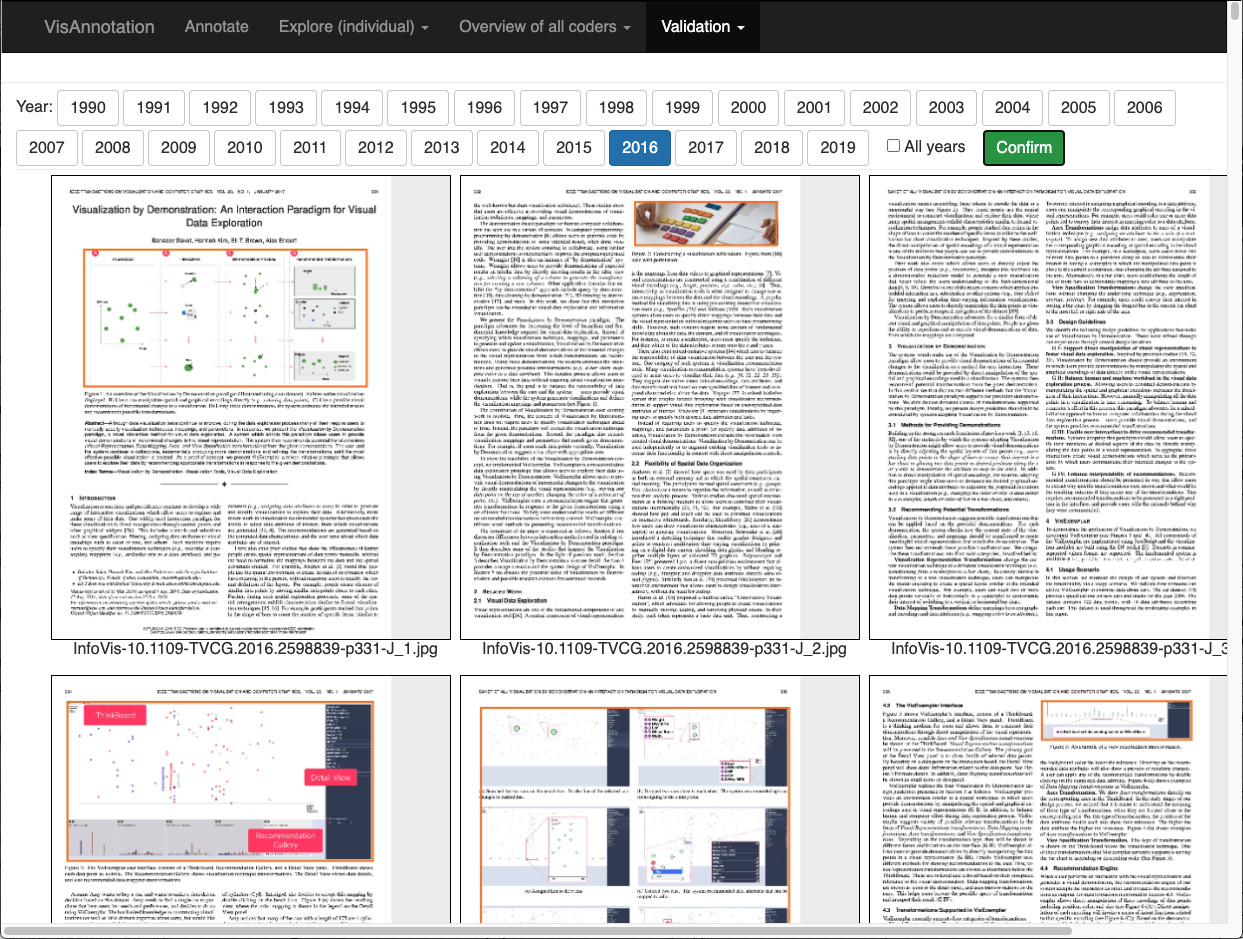}%
\label{fig:cleaningOverview}}\vspace{-1ex}
\caption{Screenshots of the tools used to clean the results of CNN-based labels. Although large labeled databases in natural scenes are becoming standard, they are comparatively rare in scholarly document databases. This tool is used for easily correcting, adding, and annotating figures and tables using the \textcolor{orange}{orange} bounding boxes and tags.}
\label{fig:cleaningTools}
\end{figure*}

\autoref{fig:cleaningTools} shows screenshots of two tools we used during our interactive cleaning process. 

\section*{C. Image Distribution in the Training Data Corpus}
\label{sec:add_c}

\changes{The diversity of image data in the training data is critical for developing useful image data extraction algorithms. We employed two databases collected by Borkin et al.~\cite{borkin2013makes} and Li and Chen~\cite{li2018toward}. We removed images if they contained abundant text. We construct an image dataset with 12 categories shown in \autoref{table:imageDistirbution}.}

\begin{table*}[!thp]
% increase table row spacing, adjust to taste
%\renewcommand{\arraystretch}{1.3}
% if using array.sty, it might be a good idea to tweak the value of
% \extrarowheight as needed to properly center the text within the cells
\caption{The image class distribution of our training data corpus. Our pseudo-paper page composer randomly pastes 10 classes of \textit{figures} and one \textit{table}, and one \textit{ bullets and equations} classes onto white paper pages. These images are subset of figures and tables from the MASSVIS and scientific visualization images collected by Borkin et al.~\cite{borkin2013makes} and Li and Chen~\cite{li2018toward}.}
\label{table:imageDistirbution} 
\centering
%% Some packages, such as MDW tools, offer better commands for making tables
%% than the plain LaTeX2e tabular which is used here.
\begin{tabular}{|l|l|l|l|l|l|}
%\hline
% & & Accuracy & Time \\
\hline
Table & Area and circles & Bars  &
Bullets and equations & Line chart  & Maps  \\
\hline
232 & 148 & 362 & 380 & 330 & 268 \\
\hline
Matrix and parallel coordinates & Multiple types & Photos &
Point-based & Scientific data visualization & Tree and Networks \\
\hline
62 & 460 & 120 & 120 & 262 & 134 \\
\hline
\end{tabular}
%\vspace{20em}
\end{table*}

%\begin{comment}
%
%\section*{C. Additional Specific \toolnameshort Use Case}
%Let us consider a neurologist in search of brain connectivity images associated with tensor field data. Using a conventional approach, he or she would retrieve papers from the databases, then inspect the titles and download and read each one-by-one. This process is cumbersome, tedious, and may break the scientist's train of thought.
 %Our neurologist may thus do a keyword search for ``diffusion tensor,'' which directly returns all figures and possible visualizations that he or she could investigate further.
 %%\ms{again, it seems to be inconsistent. In the beginning and throughout the paper, we seem to mostly talk about the idea of image and table extraction alone. But then every now and then there pops up the idea of further classifying the images into categories (i.e. analysis paper). I'm fine either way, but we should be consistent. }
 %For example, the tool returns a number of different designs for tensor fields such as multiglyphs, and the neurologist can gauge and perform non-trivial analyses across figures. He or she could, for instance, compare glyph-based, line-based, and illustrative visualizations. 
%\end{comment}

%\section*{C. Number of Case Studies By Year}

%\todo[inline]{this should say figure, but it says section} 
\section*{\changesR{D. Additional Use-Cases}}
\label{sup:cases}

\begin{figure*}[!t]
\centering
%\subfloat[Image-centric view using a ``brick-wall'' %layout.]{\includegraphics[width=.9\textwidth]{Figures/ToolUI_cropped.jpg}
%\label{fig:imageView1}}\\[1ex]%
%\subfloat[Paper-centric view using a list layout that adds title, authors, and keywords.]{\includegraphics[width=.9\textwidth]{Figures/ToolGUIList_cropped.jpg}
%\label{fig:imageView1}}\\[1ex]%
%\subfigure[Image-centric view using a ``brick wall'' layout.]{\label{fig:imageView}
%\includegraphics[width=\columnwidth]{Figures/ToolUI2.jpg}}\\[1ex]%
\includegraphics[width=\textwidth]{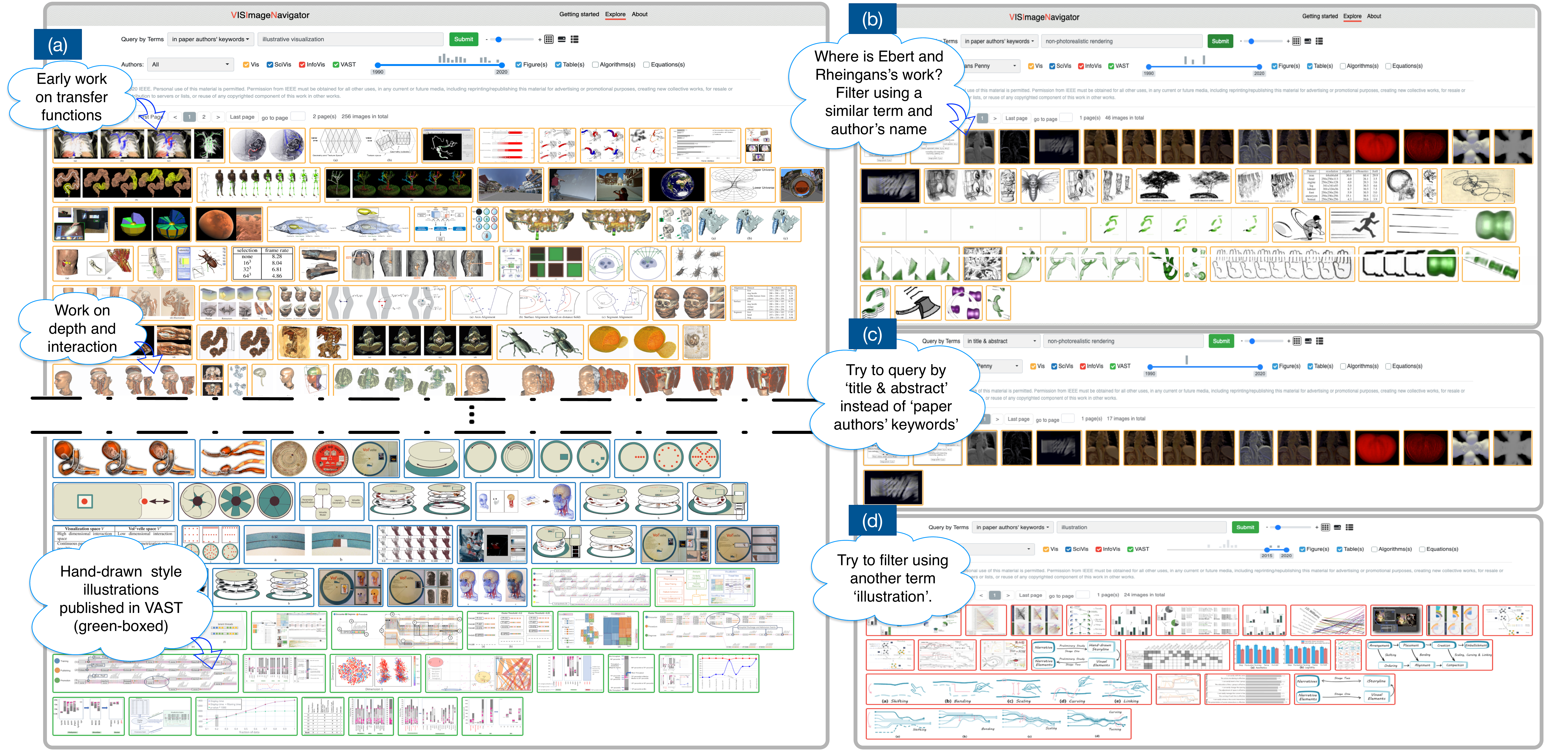}
%\\[1ex]%
\caption{\changesR{\toolnameshort use-case scenario: Identify work related to ``illustrative visualization.'' This scenario illustrates a use case when the user identifies the related work. VIN offers a focused, visual way to quickly browser related work progressively. It can complement the traditional text-based related work search.}
%The user first query ``illustrative visualization'' using the authors' keyword. The images are arranged by year and by image type (figure or table), and by conference category (Vis, SciVis, InfoVis, VAST). 
%A search by paper keywords, here ``tensor field'' returns all related figures and tables. 
}%\vspace{-1em}
\label{fig:IllustrativeVis}
\end{figure*}

\changesR{
In the paper we present three interfaces in an interactive web-based tool, VIN, that allows the general public to perform their own analyses on the entire 30 years of IEEE visualization image datasets. We show in this part example questions the VIN tool can help answer.
}

\changesR{\textbf{What are the illustrative visualization techniques in literature?}}
\changesR{Looking at images allows scholars to ``see'' the techniques invented over the years and obtain 
a gist of the development of techniques over time. \autoref{fig:IllustrativeVis} shows an exploratory process for someone, here Jerry, who has taken a visualization class in graduate school to explore techniques related to  ``illustrative visualization''. Jerry puts ``illustrative visualization'' in the query by \textit{author keywords} and ses that the tool returned 243 figures and 13 table images ranging from see-through views to distorted rendering to depth-based techniques. Jerry first observes that most of the early techniques were about transfer functions and then newer techniques address interactive exploration and augmented depth perception (\autoref{fig:IllustrativeVis}(a)--(b)). 
The most recent work in 2019 is different enough to catch Jerry's  attention in  that only  one paper at InfoVis was about non-spatial data  (\autoref{fig:IllustrativeVis}(a)). Wondering what that paper is about, Jerry switches to the paper-centric view to learn that it was
about setting parameters in a high-dimensional space. Apparently this paper has a novel use of terminology compared to other papers, which largely focused on spatial data representation. Curious about why he did not see Ebert  and Rheingans'  work on shading-based illustrative visualization,  Jerry tries a similar term `non-photorealistic  rendering'  and added  `Penny  Rheingans'  to the author’s name field; now it returned several of Rheingans' and her students' work (\autoref{fig:IllustrativeVis}(b)). Trying the same keywords and author’s field and filtering by `title and abstract' instead returns fewer results (\autoref{fig:IllustrativeVis}(c)). Jerry learns that next time he should try both search options. Further exploration  by updating  the keywords  to ``illustration'' reveals that this term was largely used in hand-drawn techniques (\autoref{fig:IllustrativeVis}(d)).}

\begin{figure*}[!tp]
\centering
%\subfloat[Image-centric view using a ``brick-wall'' %layout.]{\includegraphics[width=.9\textwidth]{Figures/ToolUI_cropped.jpg}
%\label{fig:imageView1}}\\[1ex]%
%\subfloat[Paper-centric view using a list layout that adds title, authors, and keywords.]{\includegraphics[width=.9\textwidth]{Figures/ToolGUIList_cropped.jpg}
%\label{fig:imageView1}}\\[1ex]%
%\subfigure[Image-centric view using a ``brick wall'' layout.]{\label{fig:imageView}
%\includegraphics[width=\columnwidth]{Figures/ToolUI2.jpg}}\\[1ex]%
\includegraphics[width=0.95\textwidth]{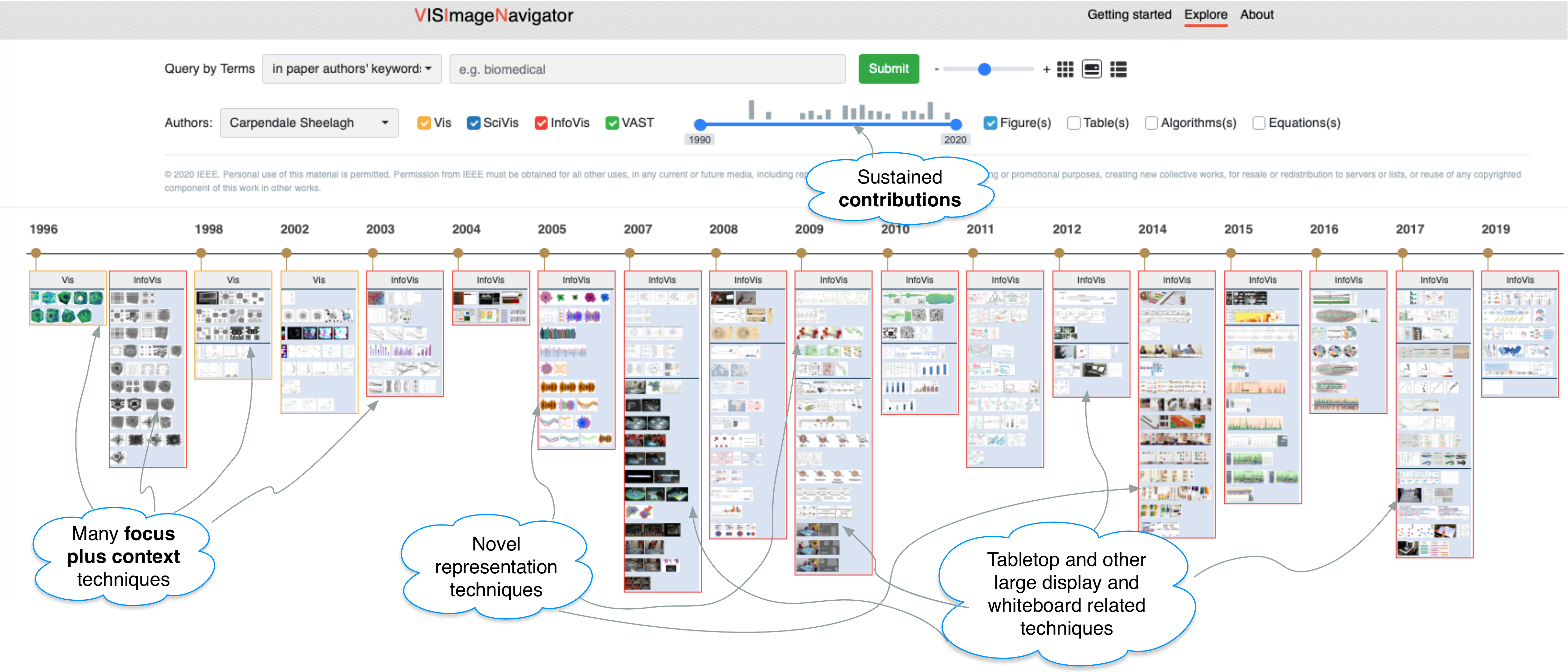}
%\\[1ex]%
\vspace{-1em}
\changesR{\caption{\toolnameshort use-case scenario: Students searching for scholarly work by Prof.\ Sheelagh Carpendale at IEEE Vis and InfoVis. This scenario describes how VIN can facilitate learning and communication. It shows the types of representations published by a specific scholar in VIS.}
%The user first query ``illustrative visualization'' using the authors' keyword. The images are arranged by year and by image type (figure or table), and by conference category (Vis, SciVis, InfoVis, VAST). 
%A search by paper keywords, here ``tensor field'' returns all related figures and tables. 
}%\vspace{-1em}
\label{fig:carpendale}
\end{figure*}

\changesR{
\textbf{What work has been published by Sheelagh Carpendale?}
Assume that Andr{\'e} is a new PhD student working with Prof.\ Carpendale at Simon Fraser University  who would  like a quick understanding of Prof.\ Carpendale's work in interactive visualization before reading her other HCI (human-computer interaction) journal and conference papers. Andr{\'e} selects the author's name from the author category and then clicks the timeline view to obtain an overview of the work (\autoref{fig:carpendale}). He sees diverse contributions in visual representations and novel interaction techniques on tabletop and large displays. 
Since Andr{\'e} likes interactive techniques, Andr{\'e} quickly begins to explore the early papers related to focus$+$context. Curious what concepts these focus plus context describe, Andr{\'e} switches to the paper view. Here he learns that Dr.\ Carpendale published several focus$+$context applications in the area of biology. 
%The interface also lets him learn ``what has Carpendale written in a specific area?'': Andr{\'e} discovers that the author made significant contributions to morphing and occlusion reduction.
Interested, Andr{\'e} decides that he should 
%also 
read these papers to learn about the details. 
%other vision and image processing works by Pfister.
}

\begin{figure*}[!htp]
\centering
\subfigure[]{\includegraphics[height=0.94\columnwidth]{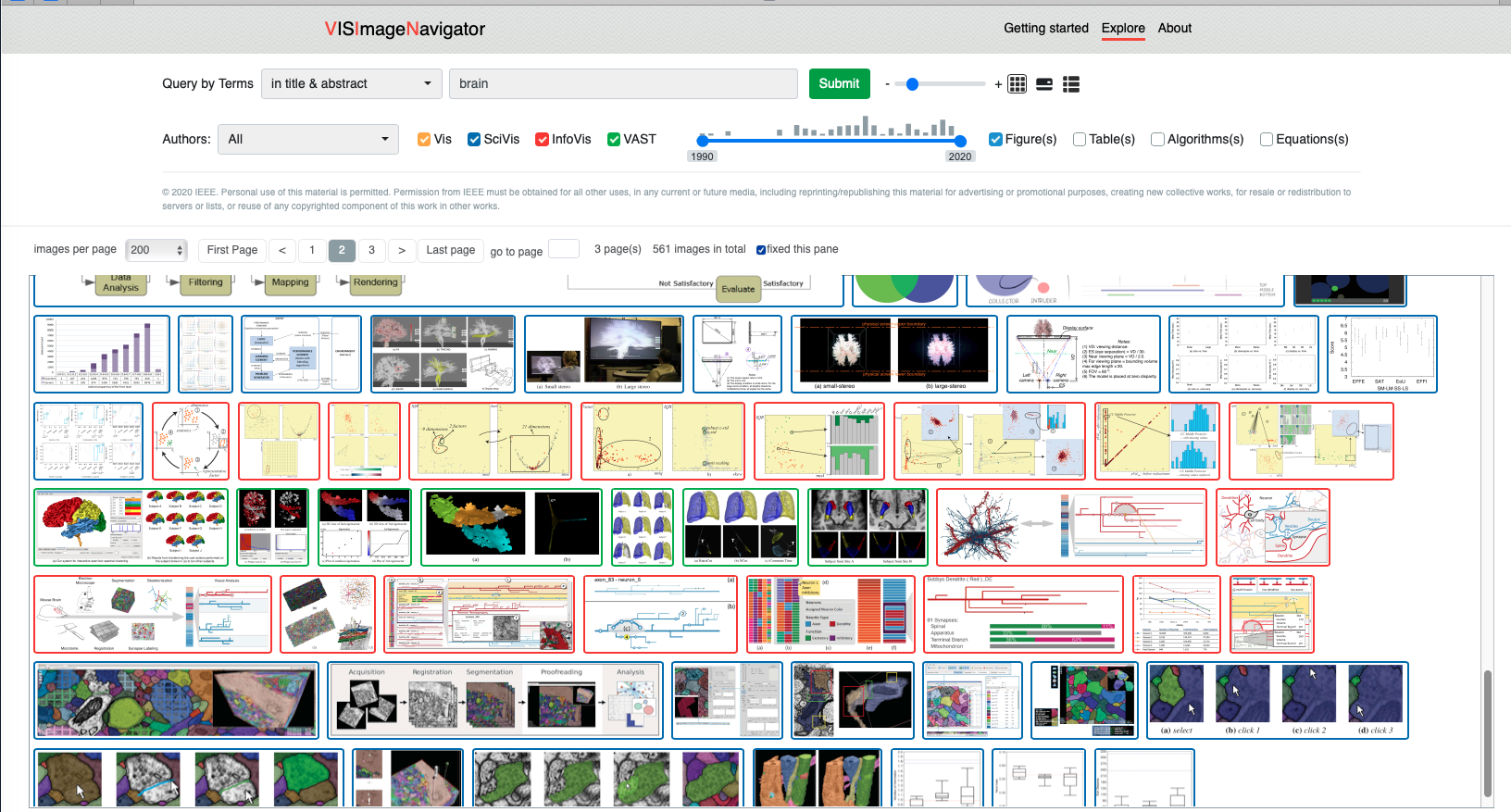}
\label{fig:brain}}\\[1ex]
%\\[1ex]%
\subfigure[]{\includegraphics[height=0.94\columnwidth]{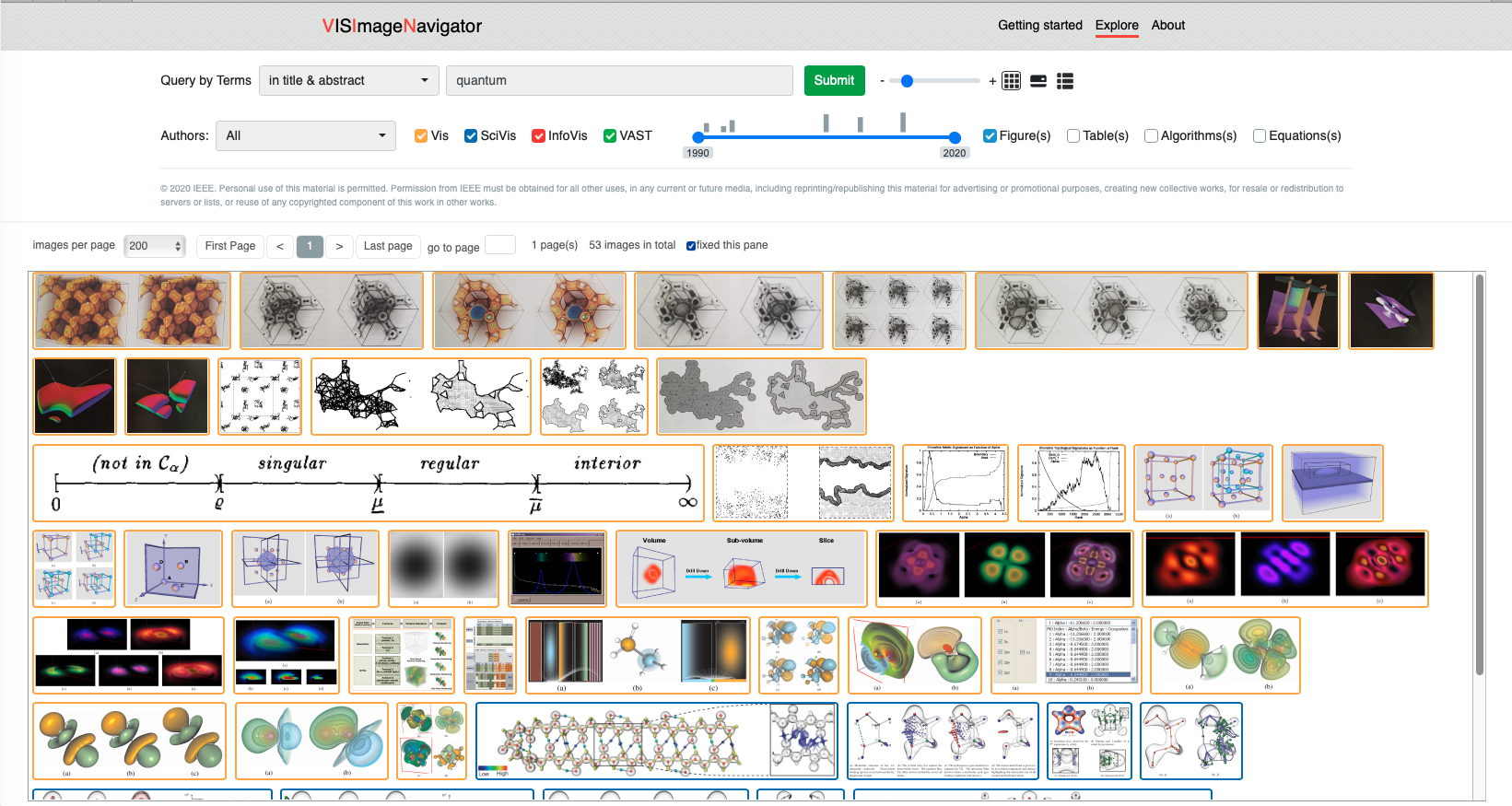}%
\label{fig:quanum}}%\vspace{-1ex}
\changesR{\caption{\toolnameshort use-case scenario: Paper images containing ``brain'' or ``quantum'' in the paper title or abstract. We see significant advances in brain visualization compared to those for exploring quantum data.}}
%}
\label{fig:app}
\vspace{-1em}
\end{figure*}

\changesR{
\textbf{What visualization applications center around quantum physics?} For Emma, a scholar whose goals are related to examine complex structures in quantum physics data,  our VIN tool is a resource for research techniques she could adapt or reuse rather than reinventing the wheel. Emma knows that she could query ``quantum'' from the authors’ keywords or more specific terms in abstract and title that target specific design choices (\eg, showing the data with line or volume rendering) or stylistic decisions (\eg, color space and line styles) (Fig.~\ref{fig:app}). 
%She knows that visualization researchers were also interested in analyzing 
She is interested in the most frequent visual encodings (e.g.,  what data attributes
are mapped to which marks and channels) and best practices 
(e.g. the use of transfer functions in volume graphics).
%She knew that visualization researchers were also interested in analyzing the most frequently used design patterns (\eg, what data attributes are mapped to what is most commonly used to encode data) or how designers reveal complex surfaces
%go against best practices 
%(\eg, 
%control the total number of colors and 
%choose appropriate transfer functions to achieve the visual effects). 
She finds that the brick wall interface provides a better understanding of the design patterns, and shows which techniques are most frequently used to show topological structures. She understands that she could create new methods to  meet the data exploration goals of the quantum physicists' new design goals.
}

%The timeline view shows that the application area has spanned the entire history of visualization with a spike around 2009 and another in 2018.

\end{multicols}

%\end{multicols}

% that's all folks
\end{document}